\documentclass[10pt]{article}
\usepackage[utf8]{inputenc}
\usepackage[margin=1in]{geometry}
\usepackage{rotating}
\usepackage{booktabs}
\usepackage{array}
\usepackage{amsmath, amsfonts, amssymb}
\usepackage{color}
\usepackage{hyperref}
\usepackage{bbm}
\usepackage{algorithm}
\usepackage[noend]{algpseudocode}
\usepackage{subcaption}
\usepackage{graphbox}
\usepackage{accents}

\usepackage{mathtools}
\mathtoolsset{showonlyrefs}

\usepackage[
    backend=biber,
    style=ieee,
    citestyle=numeric-comp,
    sorting=none,
    giveninits=true,
    natbib,
    hyperref,
    maxbibnames=99,
    doi=false,isbn=false,url=false,eprint=false
]{biblatex}

\usepackage{amsthm}
\usepackage[normalem]{ulem}
\usepackage{soul}

\newtheorem{theorem}{Theorem} 
 
\newtheorem{prop}{Proposition}
\newtheorem{cor}{Corollary} %[section]

\usepackage{chngcntr}
\usepackage{apptools}
\AtAppendix{\counterwithin{theorem}{section}}
\AtAppendix{\counterwithin{prop}{section}}
\AtAppendix{\counterwithin{cor}{section}}
\AtAppendix{\counterwithin{lemma}{section}}

\usepackage{mathtools}

\usepackage[dvipsnames]{xcolor}
\usepackage[framemethod=TikZ]{mdframed}
\usepackage{multicol}
\mdfdefinestyle{MyFrame}{%
    linecolor=black,
    outerlinewidth=0.5pt,
    innertopmargin=4pt,
    innerbottommargin=4pt,
    innerrightmargin=2pt,
    innerleftmargin=2pt,
        leftmargin = 0pt,
        rightmargin = 0pt
}

\def\Ical{\mathcal{I}}

\def\R{\mathbb{R}}
\def\E{\mathbb{E}}

\def\N{\mathbb{N}}

\newcommand{\Y}{\mathcal{Y}}
\newcommand{\X}{\mathcal{X}}
\newcommand{\K}{\mathcal{K}}

\newcommand{\C}{\mathcal{C}}
\newcommand{\T}{\mathcal{T}}

\newcommand{\ind}[1]{\mathbbm{1}\left\{#1\right\}}

\newcommand{\cR}{\mathcal{R}}

\renewcommand{\P}{\mathbb{P}}

\newcommand{\Q}{\mathbb{Q}}

\newcommand{\Var}{\mathrm{Var}}

\newcommand{\tX}{\widetilde{X}}
\newcommand{\tY}{\widetilde{Y}}

\newcommand{\Xa}{X}
\newcommand{\Xt}{\widetilde{X}}
\newcommand{\Ya}{Y}
\newcommand{\Yt}{\widetilde{Y}}

\newcommand{\thetaclass}{\hat{\theta}^{\mathrm{class}}}
\newcommand{\thetaf}{\tilde{\theta}^{f}}
\newcommand{\thetaflimit}{\theta^f}
\newcommand{\thetaPP}{\hat{\theta}^{\mathrm{PP}}}
\newcommand{\rec}{\mathbf{\Delta}}
\newcommand{\recw}{\mathbf{\Delta}^{w}}
\newcommand{\rechat}{\hat{\mathbf{\Delta}}}

\newcommand{\CPP}{\mathcal{C}^{\mathrm{PP}}}

\newcommand{\jupyter}[1]{\href{#1}{\begingroup
\setbox0=\hbox{\includegraphics[height=1.5em]{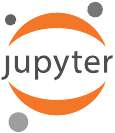}}%
\parbox{\wd0}{\box0}\endgroup}}

\DeclareMathOperator*{\argmin}{arg\,min}

\makeatletter
\def\blfootnote{\xdef\@thefnmark{}\@footnotetext}
\makeatother

\title{Prediction-Powered Inference}
\author{Anastasios N. Angelopoulos$^*$\quad Stephen Bates$^*$\quad Clara Fannjiang$^*$\\
Michael I.~Jordan$^*$\quad Tijana Zrnic$^*$\\ \\
University of California, Berkeley
}
\date{}

\begin{document}

\maketitle

\def\thefootnote{*}\footnotetext{Authors listed alphabetically. Contact:\{angelopoulos,~stephenbates,~clarafy,~michael\_jordan,~tijana.zrnic\}@berkeley.edu}
\def\thefootnote{\arabic{footnote}}

\begin{abstract}
Prediction-powered inference is a framework for performing valid statistical inference when an experimental dataset is supplemented with predictions from a machine-learning system. The framework yields simple algorithms for computing provably valid confidence intervals for quantities such as means, quantiles, and linear and logistic regression coefficients, without making any assumptions on the machine-learning algorithm that supplies the predictions. Furthermore, more accurate predictions translate to smaller confidence intervals. Prediction-powered inference could enable researchers to draw valid and more data-efficient conclusions using machine learning. The benefits of prediction-powered inference are demonstrated with datasets from proteomics, astronomy, genomics, remote sensing, census analysis, and ecology.
\end{abstract}

\section{Introduction}
\label{sec:introduction}

Imagine a scientist has a machine-learning system that can supply accurate predictions about a phenomenon far more cheaply than any gold-standard experimental technique. The scientist may wish to use these predictions as evidence in drawing scientific conclusions. For example, accurate predictions of three-dimensional structures have been made for a vast catalog of known protein sequences \cite{jumper2021highly,tunyasuvunakool2021highly} and are now being used in proteomics studies \cite{bludau2022structural,barrio2023clustering}. Such machine-learning systems are increasingly common in modern scientific inquiry, in domains ranging from cancer prognosis to microclimate modeling.  Predictions are not perfect, however, which may lead to incorrect conclusions.  Moreover, as predictions beget other predictions, these imperfections may cumulatively amplify.  How can modern science leverage machine-learning predictions in a statistically principled way? 

One way to use predictions is to follow the imputation approach: proceed as if they are gold-standard measurements. Although this lets the scientist draw conclusions cheaply and quickly due to the high-throughput nature of the machine-learning system, the conclusions may be invalid because the predictions may have biases.

Another approach is to apply the classical approach: ignore the machine-learning predictions and only use the available gold-standard measurements, which are typically far less abundant than predictions. The resulting discoveries will be statistically valid, but the smaller amount of data will limit the scope of possible discoveries.

This manuscript presents \emph{prediction-powered inference}, a framework that achieves the best of both worlds: extracting information from the predictions of a high-throughput machine-learning system, while guaranteeing statistical validity of the resulting conclusions. Prediction-powered inference provides a protocol for combining predictions, which are abundant but not always trustworthy, with gold-standard data, which is trusted but scarce, to compute confidence intervals and p-values. The resulting confidence intervals and p-values are statistically valid, as in the classical approach, but also leverage the information contained in the predictions, as in the imputation approach, to make the confidence intervals smaller and the p-values more powerful. 

Prediction-powered inference can be used with any machine-learning system. As such, it absolves the need for case-by-case analyses dependent on the machine-learning algorithm on hand. The proposed protocol thereby enables researchers to report and assess the evidence for their conclusions in a fully standardized way.

% We present \emph{prediction-powered inference}, a framework that provides an affirmative answer to the question of whether predictions can improve inferential quality.
% Rather than using predictions as raw data, prediction-powered inference uses the labeled data to estimate a mathematical object that we refer to as the \emph{rectifier}. 
% The rectifier makes it possible to transform parameter estimates based on predictions into a statistically valid confidence set.  The effectiveness of prediction-powered inference can be seen in both Figure~\ref{fig:remote-sensing} and Figure~\ref{fig:alphafold-teaser}, where the resulting confidence intervals (in green) not only cover the truth, but also are smaller than those obtained using the classical approach.
% % Notably, in Figure~\ref{fig:alphafold-teaser}, the prediction-powered interval does not contain the value one, permitting a qualitative change in the resulting scientific conclusion.

\subsection{General principle}
\label{subsec:principle}

We now overview prediction-powered inference.
The goal is to estimate a quantity $\theta^*$, such as the mean or median value of a random outcome over a population of interest. 
Towards this goal, we have access to a small gold-standard dataset of paired features and outcomes, $(X,Y) = \big( (X_1,Y_1), \ldots, (X_n, Y_n) \big)$, as well as the features from a large unlabeled dataset, $(\tX, \tY) = \big( (\tX_1, \tY_1), \ldots, (\tX_N, \tY_N) \big)$, where we do not observe the true outcomes $\tY_1,\dots,\tY_N$. 
We care about the case where $N \gg n$.
For both datasets, we have predictions of the outcome made by a machine-learning algorithm $f$, denoted $f(X) = (f(X_1),\dots,f(X_n))$ and $f(\Xt) = (f(\Xt_1),\dots,f(\Xt_N))$.

Prediction-powered inference builds confidence intervals that are guaranteed to contain $\theta^*$.
Imagine we have an estimator $\hat{\theta}$ of~$\theta^*$. 
One feasible but naive way to estimate $\theta^*$, which we call the imputation approach, is to treat the predictions as gold-standard outcomes and compute $\thetaf = \hat \theta(\Xt,f(\Xt))$. 
If the predictions are accurate, meaning $f(\Xt_i)\approx \Yt_i$, then $\thetaf$ is close to $\theta^*$.
However, $\thetaf$ will generally be biased due to errors in the predictions.
Instead, our key idea is to use the gold-standard dataset to quantify how the prediction errors affect the imputed estimate, and then construct a confidence set for $\theta^*$ by adjusting for this effect.

More systematically, the first step is to introduce a problem-specific measure of prediction error called the \emph{rectifier}, denoted as $\rec$.
The rectifier captures how errors in the predictions lead to bias in $\thetaf$. 
Intuitively, $\rec$ recovers $\theta^*$ by ``rectifying'' $\thetaf$.
The appropriate rectifier depends on the estimand of interest $\theta^*$, and we show how to derive it for a broad class of estimands.
Next, we use the gold-standard data to construct a confidence set for the rectifier, $\mathcal{R}$.
Finally, we form a confidence set for $\theta^*$ by taking $\thetaf$ and rectifying it with each possible value in the set $\mathcal{R}$.
The collection of these rectified values is the prediction-powered confidence set, $\mathcal{C}^{\rm PP}$, which is guaranteed to contain $\theta^*$ with high probability.

\begin{figure}[H]
\centering
  \begin{mdframed}[style=MyFrame,nobreak=true]
  \begin{center}
  \textbf{Prediction-Powered Inference}
  \vspace{0.3cm}
  \end{center}
  \hspace{0.2cm}
  \begin{center}
      \includegraphics[width=\linewidth]{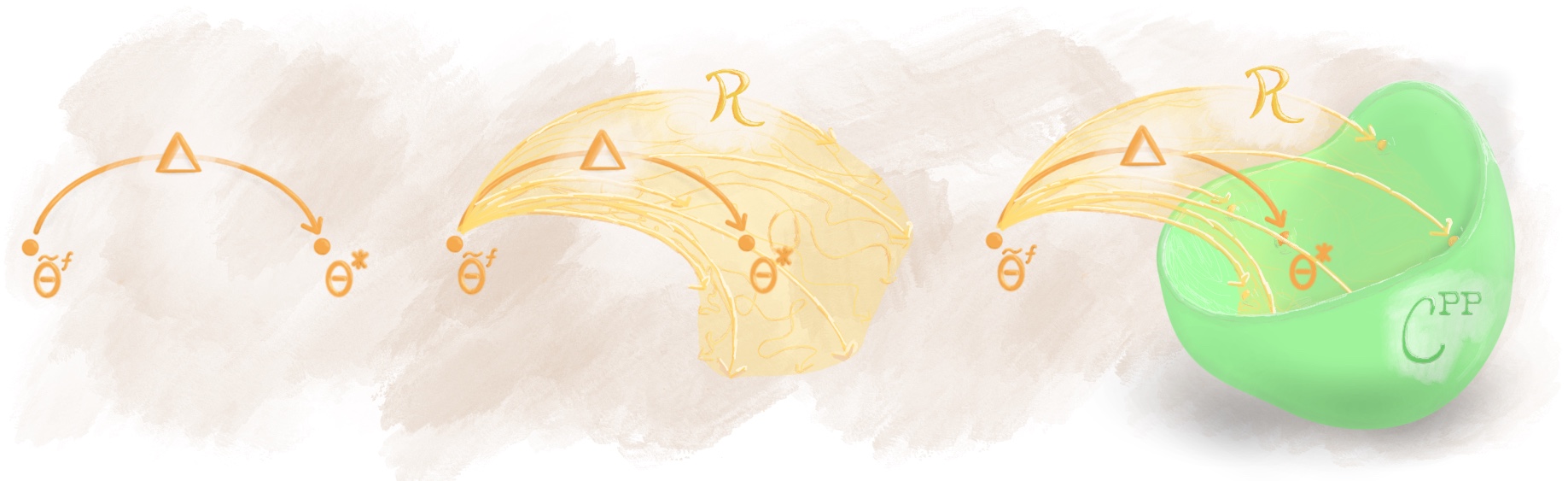}
  \end{center}
  \begin{minipage}[c]{0.26\linewidth}
      \begin{center}
      \textbf{1. Define rectifier }
      \end{center}
      \begin{center}
        Define the \emph{rectifier}, $\rec$, a measure of prediction error.
      \end{center}
\end{minipage}
\hspace{0.1cm}
\begin{minipage}[c]{0.3\linewidth}
    \begin{center}
      \textbf{2. Rectifier confidence set}
    \end{center}
    \begin{center}
      With labeled data, create $\cR$, a confidence set for the rectifier.
    \end{center}
\end{minipage}
\hspace{0.1cm}
\begin{minipage}[c]{0.4\linewidth}
      \begin{center}
        \textbf{3. Prediction-powered confidence set}
      \end{center}
      \begin{center}
        Construct confidence set $\CPP$ by \\rectifying $\thetaf$ with each value in $\cR$.
      \end{center}
\end{minipage}
  \end{mdframed}
\end{figure}
Prediction-powered inference leads to powerful and provably valid confidence intervals and p-values for a broad class of statistical problems, enabling researchers to reliably incorporate machine learning into their analyses.
We provide practical algorithms for constructing prediction-powered confidence intervals for means, quantiles, modes, linear and logistic regression coefficients, as well as other inferential targets.
For conciseness, our technical statements and algorithms will focus on constructing confidence intervals; however, note that through the duality between confidence intervals and hypothesis tests, our intervals directly imply valid prediction-powered p-values and hypothesis tests as well.

\subsection{Further preliminaries}

We use $(X,Y)\in (\X\times \Y)^n$ to denote the labeled dataset, where $X = (X_1,\dots,X_n)$ and $Y = (Y_1,\dots,Y_n)$. We use the terms ``labeled'' and ``gold-standard'' interchangeably.
We use analogous notation for the unlabeled dataset, $(\Xt,\Yt)\in (\X\times\Y)^N$, where the outcomes $\Yt$ are not observed.
For now we assume that $(X,Y)$ and $(\Xt,\Yt)$ are independently and identically distributed samples from a common distribution, $\P$. We generalize our results to settings with distribution shift and finite populations in Section \ref{sec:dist-shift} and Appendix~\ref{app:finite-pop}, respectively.
By $\theta^*$ we denote the estimand of interest, which will typically be an underlying property of $\P$, such as the mean outcome.

Next, we have a prediction rule, $f: \mathcal{X} \rightarrow \mathcal{Y}$, that is independent of the observed data. For example, it may have been trained on other data independent from both the labeled and the unlabeled data.
Thus, $f(X_i)$ denote the predictions for the labeled data and $f(\Xt_i)$ denote the predictions for the unlabeled data.
We let $f(X) = (f(X_1),\dots,f(X_n))$ and $f(\Xt) = (f(\Xt_1),\dots,f(\Xt_N))$.
We will treat $X,Y,\Xt,\Yt,f(X),f(\tX)$ as vectors and matrices where appropriate.

Our key conceptual innovation is the \emph{rectifier} $\rec$: a measure of the prediction rule's error. We formally define the rectifier in Section~\ref{sec:convex}. We use $\rechat$ to denote an estimate of the rectifier based on labeled data, which we call the empirical rectifier.

\subsection{Warmup: Mean estimation}
\label{subsec:warmup}

Before presenting our main results, we use the example of mean estimation to build intuition.
Our goal is to give a valid confidence interval for the average outcome, $\theta^* = \E[Y_i]$.  
The classical estimate of $\theta^*$ is the sample average of the outcomes on the labeled dataset, $\thetaclass = \frac{1}{n} \sum_{i=1}^n Y_i$.
We construct a prediction-powered estimate, $\thetaPP$, and show that it leads to tighter confidence intervals than $\thetaclass$ if the prediction rule is accurate.
Consider
\begin{equation}
\label{eq:PP_estimator_means}
\thetaPP = \underbrace{\frac{1}{N} \sum_{i=1}^N f(\Xt_i)}_{\thetaf} - \underbrace{\frac{1}{n} \sum_{i=1}^n (f(X_i) - Y_i)}_{\rechat}.
\end{equation}
The key idea is that if the predictions are accurate, we have $\rechat \approx 0$ and $\thetaPP \approx \frac{1}{N}\sum_{i=1}^N \Yt_i$, which has a much lower variance than $\thetaclass$ since $N\gg n$. 

Notice $\thetaPP$ is unbiased for $\theta^*$ and it is a sum of two independent terms. Thus, we can construct 95\% confidence intervals for $\theta^*$ as
\begin{equation}
    \label{eq:warmup-intervals}
    \underbrace{\thetaPP \pm 1.96 \sqrt{\frac{\hat{\sigma}^2_{f-Y}}{n} + \frac{\hat{\sigma}^2_{f}}{N}}}_{\text{prediction-powered interval}}
     \qquad \text{ or } \qquad  
     \underbrace{\thetaclass \pm 1.96 \sqrt{\frac{\hat{\sigma}^2_Y}{n}}}_{\text{classical interval}},
\end{equation}
where $\hat{\sigma}_Y^2$, $\hat{\sigma}^2_{f-Y}$, and $\hat{\sigma}^2_{f}$ are the estimated variances of the $Y_i$, $f(X_i)-Y_i$, and $f(\Xt_i)$, respectively. 
The prediction-powered confidence interval 
is smaller than the classical interval when the model is good.
Because $N \gg n$, the width of the prediction-powered interval is primarily determined by the term $\hat{\sigma}^2_{f-Y}$.
Furthermore, when the model has small errors, we have $\hat{\sigma}^2_{f-Y} \ll \hat{\sigma}^2_{Y}$. 
Thus, the width of the prediction-powered interval will be smaller than the width of the classical interval. This estimator exists in many forms in the literature---see Section~\ref{sec:lit_review}.
This variance reduction is why prediction-powered confidence intervals are smaller than their classical counterparts in a broad range of settings beyond mean estimation.

\subsection{Related work}
\label{sec:lit_review}

Our technical results generalize tools from the model-assisted survey sampling literature~\cite[e.g.,][]{sarndal2003model}, which provides methods to improve inference from surveys in the presence of auxiliary information.
In particular, the mean estimator in Section~\ref{subsec:warmup} is the difference estimator, closely related to generalized regression estimators~\cite{cassel1976some}.
It has long been recognized that model predictions can be leveraged as auxiliary data~\cite{wu2001model}, and much work has gone into producing asymptotically valid confidence intervals when the predictive model is fit on the same data that is used for inference---see~\cite{breidt2017model} for a recent overview. Our work is also related to the statistical literature on semiparametric inference, missing data, and multiple imputation~\citep[e.g.,][]{little2019statistical}. In particular, Robins et al.~\cite{robins1994estimation}, Robins and Rotnitzky ~\cite{robins1995semiparametric}, Chen and Breslow~\cite{chen2004semiparametric}, Yu and Nan \cite{yu2006revisit} study regression with missing data.
The rectifier resembles debiasing strategies that are pervasive in this literature, an example being the AIPW estimator~\cite{robins1995semiparametric}.
Likewise, our setting is related to measurement error~\citep[e.g.,][]{carroll2006measurement}, particularly to Chen et al.~\cite{chen2005measurement}, who study the estimation of parameters defined as solutions to many estimating equations, as we will in this work.
Prediction-powered inference aims to provide simple, broadly applicable algorithms using similar debiasing tricks, while allowing the use of state-of-the-art black-box machine-learning systems.

Recently, a body of work on estimation with many unlabeled data points and few labeled data points has been developed~\citep{pepe1992inference, chen2003information, wasserman2007, zhang2019semisupervised, azriel2022semisupervised, song2023general}, focusing on efficiency in semiparametric or high-dimensional regimes.
In particular, Chakrabortty and Cai~\cite{Chakrabortty2018} study efficient estimation of linear regression parameters, Chakrabortty et al.~\cite{chakrabortty2022semi, chakrabortty2022general} study efficient quantile estimation and quantile treatment effect estimation with high-dimensional covariates, Zhang and Bradic~\cite{zhang2022high} study mean estimation in a high-dimensional setting, Deng et al.~\cite{deng2020optimal} study linear regression parameters in a high-dimensional setting, and Hou et al.~\cite{hou2021surrogate} study an imputation approach to improving generalized linear models.
Finally, Song et al.~\cite{song2023general} study M-estimation, using a projection-based correction to the classical M-estimator loss based on simple statistics (e.g. low-order polynomials) of the features.
Prediction-powered inference continues in this vein but focuses on the setting where the scientist has access to a good predictive model fit on separate data and makes no assumptions about the model (such as consistency).
The confidence intervals and resulting p-values from previous work rely on asymptotic approximations, while prediction-powered inference has both asymptotic and nonasymptotic variants.
Furthermore, prediction-powered inference goes beyond random sampling and considers certain forms of distribution shift.

More distantly, our setting, in which we have access to some labeled data alongside unlabeled data, also appears in semisupervised learning~\citep[e.g.,][]{zhu2005semi,zhu2009introduction}, which studies the question of how to improve prediction accuracy with unlabeled data. 
We also refer the reader to the related literatures on transfer learning~\citep[e.g.,][]{bastani2021predicting, xu2021learning, tian2022transfer, lin2022transfer} and surrogates in causal inference~\citep[e.g.,][]{kallus2020role}.
Thematically, our work is most similar to the work of Wang et al.~\cite{wang2020methods}, who also introduce a method to correct machine-learning predictions for the purpose of subsequent inference. However, our work provides confidence intervals that are provably valid under minimal assumptions about the data-generating distribution, whereas Wang et al.\ require certain parametric assumptions about the relationship between the prediction model and the true response.
We compare against this baseline in Appendix~\ref{app:comparisons}.

\section{Main theory: Convex estimation}
\label{sec:convex}

Our main contribution is a technique for inference on estimands that can be expressed as the solution to a \emph{convex optimization problem}.
In addition to means, this includes medians, other quantiles, linear and logistic regression coefficients, and many other quantities.
Formally, we consider estimands of the form
\begin{equation}
    \label{eq:convex-minimizer}
    \theta^* = \argmin_{\theta \in\R^p} \; \E\left[\ell_{\theta}(X_i, Y_i)\right],
\end{equation}
for a loss function $\ell_{\theta} : \X \times \Y \to \R$ that is convex in $\theta \in \R^p$, for some $p\in\N$.
Throughout, we take the existence of $\theta^*$ as given.
If the minimizer is not unique, our method will return a confidence set guaranteed to contain all minimizers.
Under mild conditions, convexity ensures that $\theta^*$ can also be expressed as the value solving
\begin{equation}
    \label{eq:gradient-zero}
    \E\left[g_{\theta^*}(X_i, Y_i) \right] = 0,
\end{equation}
where $g_{\theta} : \X \times \Y \to \R^p$ is a subgradient of $\ell_{\theta}$ with respect to $\theta$.
We will call convex estimation problems where $\theta^*$ satisfies \eqref{eq:gradient-zero} nondegenerate, and we will later discuss mild conditions that ensure this regularity.

\paragraph{Defining the rectifier.} Following the outline in Section~\ref{subsec:principle}, the first step in prediction-powered inference is to define a rectifier.
As in the mean estimation case, the rectifier captures a notion of prediction error. In the general setting of convex estimation problems, the relevant notion of error is the bias of the subgradient $g_{\theta}$ computed using the predictions:
\begin{equation}
\label{eq:convex-rectifier}
    \rec_{\theta} = \E \left[g_\theta(X_i, Y_i) - g_\theta(X_i, f(X_i))\right].
\end{equation}
% The connection to estimating $\theta^*$ is clear: we can decompose the gradient condition~\eqref{eq:gradient-zero} as $\E\left[g_{\theta^*}(X_i, Y_i) \right] = \E\left[g_{\theta^*}(X_i, f(X_i)) \right] + \rec_{\theta} = 0$.
% Using plug-in estimates for each term in the final equality results in a natural prediction-powered point estimate; see Section~\ref{subsec:point-estimate} for more detail.

\paragraph{Rectifier confidence set.} The second step is to create a confidence set for the rectifier, $\cR_{\delta}(\theta)$, satisfying 
\begin{equation}
\label{eq:convex-rec-set}
    P\left(\rec_{\theta} \in \cR_{\delta}(\theta) \right) \geq 1-\delta.
\end{equation}
Because the rectifier is an expectation for each $\theta$, $\cR_{\delta}(\theta)$ can be constructed using standard, off-the-shelf confidence intervals for the mean, which we review in Appendix~\ref{app:cis}.

\paragraph{Prediction-powered confidence set.} The final step is to form a confidence set for $\theta^*$.
We do so by combining $\cR_\delta(\theta)$ with a term that accounts for finite-sample fluctuations due to having $N$ unlabeled data points. In particular, for every $\theta$, we want a confidence set $\T_{\alpha-\delta}(\theta)$ for $g_{\theta}^f = \E [g_\theta (X_i,f(X_i)) ]$, satisfying
$$P\left(g_{\theta}^f \in \T_{\alpha-\delta}(\theta) \right)\geq 1 - (\alpha-\delta).$$
Again, since $g_{\theta}^f$ is a mean, constructing $\T_{\alpha-\delta}(\theta)$ is easy and can be done with off-the-shelf tools.

We put all the steps together in Theorem \ref{thm:convex-validity}.

\begin{theorem}[Convex estimation]
\label{thm:convex-validity}
Suppose that the convex estimation problem is nondegenerate as in \eqref{eq:gradient-zero}. Fix $\alpha\in(0,1)$ and $\delta\in(0,\alpha)$.
Suppose that, for any $\theta\in\R^p$, we can construct $\cR_\delta(\theta)$ and $\T_{\alpha-\delta}(\theta)$ satisfying
$$P\left(\rec_{\theta} \in \cR_{\delta}(\theta) \right) \geq 1-\delta; \quad P\left(g_{\theta}^f \in \T_{\alpha-\delta}(\theta) \right) \geq 1-(\alpha-\delta).$$
Let $\CPP_\alpha = \left\{\theta : 0 \in \cR_\delta(\theta) + \T_{\alpha-\delta}(\theta)\right\}$, where $+$ denotes the Minkowski sum.\footnote{The Minkowski sum of two sets $A$ and $B$ is equal to $\{a+b: a\in A, b\in B\}$.} Then,
\begin{equation}
    P( \theta^* \in \CPP_\alpha ) \geq 1-\alpha.
\end{equation}
\end{theorem}
This result means that we can construct a valid confidence set for $\theta^*$, without assumptions about the data distribution or the machine-learning model, for any nondegenerate convex estimation problem.
We also present an asymptotic counterpart of Theorem~\ref{thm:convex-validity} in Appendix~\ref{app:asymp-convex-validity}.

Most practical problems are nondegenerate \eqref{eq:gradient-zero}. For example,
if the loss is differentiable for all $\theta\in\R^p$, then the problem is immediately nondegenerate. 
Furthermore, if the data distribution does not have point masses and, for every $\theta$, $\ell_\theta(x,y)$ is nondifferentiable only for a measure-zero set of $(x,y)$ pairs, then the problem is again nondegenerate.

We have focused on convex estimation problems, since this is a broad class of estimands addressed by prediction-powered inference. Nonetheless, we highlight that the general principles for prediction-powered inference from Section~\ref{subsec:principle} are applicable more broadly, and lead to additional results and algorithms for other estimands and some forms of distribution shift; see Section~\ref{sec:methods-details} for such extensions.

\subsection{Algorithms}
\label{subsec:explicit_algs}

In this section we present prediction-powered algorithms for several canonical inference problems.  We defer the proofs of their validity to Appendix \ref{app:proofs}. 
The algorithms rely on confidence intervals derived from the central limit theorem. We implicitly assume the standard, mild regularity conditions required for the asymptotic validity of such intervals, which we overview in Appendix \ref{app:regularity-conditions}. We also present a parallel set of algorithms that are obtained via nonasymptotic constructions in Appendix \ref{app:nonasymptotic-algos}. In the algorithms we use $z_{1-\delta}$ to denote the $1-\delta$ quantile of the standard normal distribution, for $\delta\in(0,1)$.
All algorithms are technically simplified versions of Algorithm~\ref{alg:general-convex} with different choices of gradients and rectifiers; see Table~\ref{table:grad-rec} for the correspondence.

\paragraph{Mean estimation.} We begin by returning to the problem of mean estimation:
\begin{equation}
\label{eq:mean-outcome}
\theta^* = \E[Y_i].
\end{equation}
The mean can alternatively be expressed as the solution to a convex optimization problem by writing it as the minimizer of the average squared loss:
$$\theta^* = \argmin_{\theta\in\R} \E[\ell_\theta(Y_i)] = \argmin_{\theta\in\R} \E\left[\frac{1}{2}(Y_i - \theta)^2\right].$$
The squared loss $\ell_\theta(y)$ is differentiable, with gradient equal to $g_\theta(y) = \theta - y$. Applying this in the definition of the rectifier \eqref{eq:convex-rectifier}, we get
$\rec_{\theta} \equiv \rec = \E[f(X_i) - Y_i]$.
Note that this rectifier has no dependence on $\theta$.  
We provide an explicit algorithm for prediction-powered mean estimation and its guarantee in Algorithm \ref{alg:means} and Proposition \ref{prop:mean_est_validity}, respectively.

\begin{prop}[Mean estimation]
\label{prop:mean_est_validity} 
Let $\theta^*$ be the mean outcome \eqref{eq:mean-outcome}. Then, the prediction-powered confidence interval in Algorithm~\ref{alg:means} has valid coverage: $\liminf_{n,N\rightarrow\infty}P\left(\theta^* \in  \CPP_\alpha\right) \ge 1-\alpha$.
\end{prop}

\paragraph{Quantile estimation.}
We now turn to quantile estimation.
For a pre-specified level $q\in(0,1)$, we wish to estimate the $q$-quantile of the outcome distribution:
\begin{equation}
\label{eq:q-quantile}
\theta^* = \min\left\{\theta: P\left( Y_i \leq \theta\right) \geq q\right\}.
\end{equation}
To simplify the exposition, we assume that the distribution of $Y_i$ does not have point masses; this ensures that the problem is nondegenerate \eqref{eq:gradient-zero}, though it is  possible to generalize beyond this setting with a standard construction.
It is well known~\cite{koenker1978regression} that the $q$-quantile can be expressed in variational form as
\begin{equation}
    \label{eq:quantile-def}
    \theta^* = \argmin_{\theta\in\R} \; \E\left[\ell_{\theta}(Y_i)\right] = \argmin_{\theta\in\R} \; \E\left[q(Y_i-\theta)\ind{Y_i > \theta} + (1-q)(\theta-Y_i)\ind{Y_i \leq \theta}\right],
\end{equation}
where $\ell_{\theta}$ is called the quantile loss (or ``pinball'' loss). The quantile loss has subgradient $g_{\theta}(y) = -q\ind{y > \theta} + (1-q)\ind{y \leq \theta} = -q + \ind{y \leq \theta}$.
Plugging the expression for $g_{\theta}(y)$ into the definition \eqref{eq:convex-rectifier}, we get the relevant rectifier: $\rec_{\theta} = P(Y_i\leq \theta) - P(f(X_i)\leq \theta) = \E \left[\ind{Y_i\leq \theta} - \ind{f(X_i)\leq \theta}\right]$. In Algorithm \ref{alg:quantile} we state an algorithm for prediction-powered quantile estimation; see Proposition \ref{prop:quantile} for a statement of validity.

\begin{prop}[Quantile estimation]
\label{prop:quantile}
    Let $\theta^*$ be the $q$-quantile~\eqref{eq:q-quantile}. Then, the prediction-powered confidence set in Algorithm~\ref{alg:quantile} has valid coverage: $\liminf_{n,N\rightarrow\infty} P\left(\theta^* \in \CPP_{\alpha}\right) \geq 1-\alpha$.
\end{prop}

\paragraph{Logistic regression.} In logistic regression, the target of inference is defined by
\begin{equation}
\label{eq:logistic-sol}
    \theta^* = \argmin_{\theta\in\R^d} \E[\ell_\theta(X_i,Y_i)] = \argmin_{\theta\in\R^d} \E\left[-Y_i \theta^\top X_i + \log(1 + \exp(\theta^\top X_i))\right],
\end{equation}
where $Y_i\in\{0,1\}$. The logistic loss is differentiable and hence the optimality condition \eqref{eq:gradient-zero} is ensured. Its gradient is equal to $g_{\theta}(x,y) = -x y + x \mu_{\theta}(x)$,
where $\mu_\theta(x) = 1 / (1 + \exp(-x^\top \theta))$ is the predicted mean for point $x\in\X$ based on parameter vector $\theta$. Other generalized linear models (GLMs) have the same gradient form, and thus also optimality condition~\eqref{eq:gradient-zero}, but for a different mean predictor $\mu_\theta(x)$ (see Chapter~3 of Efron~\cite{efron_2022}). For example, Poisson regression uses $\mu_\theta(x) = \exp(x^\top \theta)$. In view of our general solution for convex estimation, the rectifier is constant for all $\theta$ and equal to $\rec_{\theta} \equiv \rec = \mathbb{E}\left[X_i (f(X_i) - Y_i)\right]$.
In Algorithm \ref{alg:logistic} we state a method for prediction-powered logistic regression and in Proposition \ref{prop:logistic} we provide its guarantee. We use $X_{i,j}$ to denote the $j$-th coordinate of point $X_i$. Poisson regression is handled in essentially the same way: concretely, in Algorithm~\ref{alg:logistic} we simply change the choice of $\mu_\theta(x)$ defined in line 5.

\begin{prop}[Logistic regression]
\label{prop:logistic}
Let $\theta^*$ be the logistic regression solution \eqref{eq:logistic-sol}. Then, the prediction-powered confidence set in Algorithm~\ref{alg:logistic} has valid coverage: $\liminf_{n,N\rightarrow\infty}P\left(\theta^* \in  \CPP_\alpha\right) \ge 1-\alpha$.
\end{prop}

\paragraph{Linear regression.}
Finally, we consider inference for linear regression:
\begin{equation}
\label{eq:linear-sol}
\theta^* = \argmin_{\theta\in\R^d} \E[\ell_\theta(X_i,Y_i)] = \argmin_{\theta\in\R^d} \E [(Y_i - X_i^\top \theta)^2].
\end{equation}
While it is possible to obtain an algorithm for linear regression based on Theorem \ref{thm:convex-validity}, one can derive a more powerful solution by using the fact that the natural estimator for problem \eqref{eq:linear-sol} is linear in $Y$. We exploit these further properties in Algorithm \ref{alg:ols} and Proposition \ref{prop:ols}, where we state a method for prediction-powered linear regression and establish its validity, respectively.

\begin{prop}[Linear regression]
\label{prop:ols}
Let $\theta^*$ be the linear regression solution \eqref{eq:linear-sol} and fix $j^*\in[d]$. Then, 
the prediction-powered confidence interval in Algorithm~\ref{alg:ols} has valid coverage: $\liminf_{n,N\rightarrow\infty}P\left(\theta^*_{j^*} \in  \CPP_\alpha\right) \ge~1-~\alpha$.
\end{prop}

% \newpage

\begin{algorithm}[H]
\caption{Prediction-powered mean estimation}
\label{alg:means}
\begin{algorithmic}[1]
\Require labeled data $(\Xa, \Ya)$, unlabeled features $\Xt$, predictor $f$, error level $\alpha\in(0,1)$
\State $\thetaPP \leftarrow \thetaf - \rechat := \frac{1}{N}\sum_{i=1}^N f(\Xt_i) - \frac{1}{n}\sum_{i=1}^n (f(X_i) - Y_i)$ \Comment{prediction-powered estimator}
\State $\hat \sigma_{f}^2 \leftarrow \frac 1 N \sum_{i=1}^N (f(\Xt_i) - \thetaf)^2$ \Comment{empirical variance of imputed estimate}
\State $\hat \sigma_{f-Y}^2 \leftarrow \frac 1 n \sum_{i=1}^n (f(X_i) - Y_i - \rechat )^2$ \Comment{empirical variance of empirical rectifier}
\State $w_\alpha \leftarrow z_{1-\alpha/2} \sqrt{\frac{\hat\sigma_{f-Y}^2}{n} + \frac{\hat\sigma_{f}^2}{N}}$ \Comment{normal approximation}
\Ensure 
prediction-powered confidence set $\CPP_\alpha = \left(\thetaPP \pm w_\alpha\right)$
\end{algorithmic}
\end{algorithm}

\vspace{-.4cm}

\begin{algorithm}[H]
\caption{Prediction-powered quantile estimation}
\label{alg:quantile}
\begin{algorithmic}[1]
\Require labeled data $(\Xa, \Ya)$, unlabeled features $\Xt$, predictor $f$, quantile $q\in(0,1)$, error level $\alpha\in(0,1)$
\State Construct fine grid $\Theta_{\rm grid}$ between $\min_{i\in[N]} f(\Xt_i)$ and $\max_{i\in[N]} f(\Xt_i)$ 
\For{$\theta \in \Theta_{\rm grid}$}
\State $\rechat_{\theta} \leftarrow \frac{1}{n}\sum_{i=1}^n (\ind{Y_i \leq \theta} - \ind{f(X_i) \leq \theta})$ \Comment{empirical rectifier}
\State $\hat F(\theta)  \leftarrow \frac{1}{N}\sum_{i=1}^N  \ind{f(\Xt_i) \leq \theta}$ \Comment{imputed CDF}
\State $\hat{\sigma}_\Delta^2(\theta)  \leftarrow \frac{1}{n}\sum_{i=1}^n\left(\ind{Y_i \leq \theta} - \ind{f(X_i) \leq \theta} - \rechat_{\theta}\right)^2$ \Comment{empirical variance of empirical rectifier}
\State $\hat{\sigma}_{g}^2(\theta) \gets \frac{1}{N}\sum_{i=1}^N\left(\ind{f(\Xt_i) \leq \theta} - \hat F(\theta) \right)^2$ \Comment{empirical variance of imputed CDF}
\State $w_\alpha(
\theta) \leftarrow z_{1-\alpha/2}\sqrt{\frac{\hat{\sigma}^2_\Delta(\theta)}{n} + \frac{\hat{\sigma}^2_{g}(\theta)}{N}}$ \Comment{normal approximation}
\EndFor
\Ensure 
prediction-powered confidence set $\CPP_\alpha = \left\{ \theta \in \Theta_{\rm grid} :  |\hat F(\theta) + \rechat_{\theta} - q|  \leq w_\alpha(\theta) \right\}$
\end{algorithmic}
\end{algorithm}

\vspace{-.4cm}

\begin{algorithm}[H]
\caption{Prediction-powered logistic regression}
\label{alg:logistic}
\begin{algorithmic}[1]
\Require labeled data $(\Xa, \Ya)$, unlabeled features $\Xt$, predictor $f$, error level $\alpha \in(0,1)$ 
\State Construct fine grid $\Theta_{\rm grid} \subset \mathbb{R}^d$ of possible coefficients
\State $\rechat_j \leftarrow \frac{1}{n}\sum_{i=1}^n X_{i,j}(f(X_i) - Y_i), \quad j\in[d]$ 
\Comment{empirical rectifier}
\State $\hat \sigma_{\Delta,j}^2 \leftarrow \frac{1}{n}\sum_{i=1}^n\left(X_{i,j}(f(X_i) - Y_i) - \rechat_j\right)^2, \quad j\in[d]$ 
\Comment{empirical variance of empirical rectifier}
\For{$\theta \in \Theta_{\rm grid}$}
\State $\hat g^f_{\theta, j} \leftarrow \frac{1}{N}\sum_{i=1}^N \Xt_{i,j}\left(\mu_\theta(\Xt_i) - f(\Xt_i)\right), \quad j\in [d]$, \quad where $\mu_\theta(x) = \frac{1}{1+\exp(-x^\top \theta)}$ \Comment{imputed gradient}
\State $\hat \sigma_{g,j}^2(\theta) \leftarrow \frac{1}{N}\sum_{i=1}^N(\Xt_{i,j}(\mu_\theta(\Xt_i) - f(\Xt_i)) - \hat g_{\theta,j}^f)^2, j\in[d]$  \Comment{empirical variance of imputed gradient}
\State $w_{\alpha,j}(\theta) \leftarrow z_{1-\alpha/(2d)} \sqrt{ \frac{\hat\sigma_{\Delta,j}^2}{n} +  \frac{\hat \sigma_{g,j}^2(\theta)}{N}}, \quad j\in[d]$ \Comment{normal approximation}
\EndFor
\Ensure 
prediction-powered confidence set $\CPP_\alpha = \left\{\theta\in \Theta_{\rm grid}:  |\hat g^f_{\theta, j}  + \rechat_j| \leq w_{\alpha,j}(\theta), \forall j\in[d] \right\}$
\end{algorithmic}
\end{algorithm}

\vspace{-.4cm}

\begin{algorithm}[H]
\caption{Prediction-powered linear regression}
\label{alg:ols}
\begin{algorithmic}[1]
\Require labeled data $(X, Y)$, unlabeled features $\Xt$, predictor $f$, coefficient $j^*\in[d]$, error level $\alpha\in(0,1)$
\State $\thetaPP \leftarrow \thetaf - \rechat := \Xt^\dagger f(\Xt) - X^\dagger (f(X) - Y)$ \Comment{prediction-powered estimator}
\State $\tilde \Sigma \leftarrow \frac 1 N \Xt^\top \Xt$, $\tilde M \leftarrow \frac 1 N \sum_{i=1}^N (f(\Xt_i) - \Xt_i^\top  \thetaf)^2 \Xt_i \Xt_i^\top$
\State $\tilde V \leftarrow \tilde\Sigma^{-1}\tilde M \tilde \Sigma^{-1}$ \Comment{``sandwich'' variance estimator for imputed estimate}
\State $\Sigma \leftarrow \frac 1 n X^\top X$,  $M\leftarrow \frac 1 n \sum_{i=1}^n (f(X_i) - Y_i - X_i^\top  \rechat)^2 X_i X_i^\top$
\State $V \leftarrow \Sigma^{-1} M  \Sigma^{-1}$ \Comment{``sandwich'' variance estimator for empirical rectifier}
\State $w_\alpha \leftarrow z_{1-\alpha/2} \sqrt{ \frac{ V_{j^*j^*}}{n} + \frac{\tilde V_{j^*j^*}}{N}}$ \Comment{normal approximation}
\Ensure 
prediction-powered confidence set $\CPP_\alpha = \left(\thetaPP_{j^*} \pm w_\alpha\right)$
\end{algorithmic}
\end{algorithm}

\vspace{-.4cm}

\begin{algorithm}[H]
\caption{Prediction-powered convex estimation}
\label{alg:general-convex}
\begin{algorithmic}[1]
\Require labeled data $(\Xa, \Ya)$, unlabeled features $\Xt$, predictor $f$, error level $\alpha\in(0,1)$
\State Construct fine grid $\Theta_{\rm grid}$
\For{$\theta \in \Theta_{\rm grid}$}
\State $\rechat_{\theta, j} \leftarrow \frac{1}{n}\sum_{i=1}^n (g_\theta(X_i,Y_i)_j - g_\theta(X_i,f(X_i))_j)$ \Comment{empirical rectifier}
\State $\hat{g}_{\theta, j}^f =\gets \frac{1}{N}\sum_{i=1}^N g_\theta(\tX_i,f(\tX_i))_j$ \Comment{imputed gradient}
\State $\hat{\sigma}_\Delta^2(\theta)  \leftarrow \frac{1}{n}\sum_{i=1}^n\left(g_\theta(X_i,Y_i)_j - g_\theta(X_i,f(X_i))_j - \rechat_{\theta}\right)^2$ \Comment{empirical variance of empirical rectifier}
\State $\hat{\sigma}_{g}^2(\theta) \gets \frac{1}{N}\sum_{i=1}^N\left(g_\theta(\tX_i,f(\tX_i))_j - \hat g_{\theta, j}^f \right)^2$ \Comment{empirical variance of imputed gradient}
\State $w_\alpha(
\theta) \leftarrow z_{1-\alpha/2}\sqrt{\frac{\hat{\sigma}^2_\Delta(\theta)}{n} + \frac{\hat{\sigma}^2_{g}(\theta)}{N}}$ \Comment{normal approximation}
\EndFor
\Ensure 
prediction-powered confidence set $\CPP_\alpha = \left\{ \theta \in \Theta_{\rm grid} :  | \hat g_{\theta, j}^f + \rechat_{\theta} |  \leq w_\alpha(\theta) \right\}$
\end{algorithmic}
\end{algorithm}

\vspace{-.4cm}

\begin{table}[H]
\centering
\small
\begin{tabular}{m{2.6cm} m{5cm} m{6cm} m{1.3cm}}
\toprule
Estimand & Prediction-based gradient $\hat g_\theta^f$ & Rectifier $\rechat_\theta$ & Procedure \\
\midrule
Mean & $\theta - \frac{1}{N} \sum_{i=1}^{N} f(\tX_i)$ & $\frac{1}{n} \sum_{i=1}^{n} (f(X_i) - Y_i)$ & Alg. \ref{alg:means} \\
\addlinespace
Median & $\frac{1}{2N} \sum_{i=1}^{N} \text{sign} (\theta-f(\tX_i))$ & $\frac{1}{n} \sum_{i=1}^{n} \left( \ind{f(X_i) \leq \theta} - \ind{Y_i \leq \theta} \right)$ & Alg. \ref{alg:quantile} \\
\addlinespace
$q$-quantile & $-q + \frac{1}{N} \sum_{i=1}^{N} \ind{f(\tX_i) \leq \theta}$ & $\frac{1}{n} \sum_{i=1}^{n} \left( \ind{f(X_i) \leq \theta} - \ind{Y_i \leq \theta} \right)$ & Alg. \ref{alg:quantile} \\
\addlinespace
Logistic regression & $\frac{1}{N} \sum_{i=1}^{N} \tX_i^T \left( \frac{1}{1+e^{-\hat{\theta}^T \tX_i}} - f(\tX_i) \right)$ & $\frac{1}{n} \sum_{i=1}^{n} X_i (f(X_i) - Y_i)$ & Alg. \ref{alg:logistic} \\
\addlinespace
Linear regression & $\theta - \tX^T f(\tX)$ & $X^+(f(X) - Y)$ & Alg. \ref{alg:ols} \\
\addlinespace
Convex minimizer & $\frac{1}{N} \sum_{i=1}^{N} \nabla \ell_\theta (\tX_i, f(\tX_i))$ & $\frac{1}{n} \sum_{i=1}^{n} \left( \nabla \ell_\theta (X_i, f(X_i)) - \nabla \ell_\theta (X_i, Y_i) \right)$ & Alg. \ref{alg:general-convex} \\
\bottomrule
\end{tabular}
\caption{Prediction-powered inference for common statistical problems.}
\label{table:grad-rec}
\end{table}

\begin{figure}[H]
    \centering 
    \includegraphics[width=0.7\textwidth]{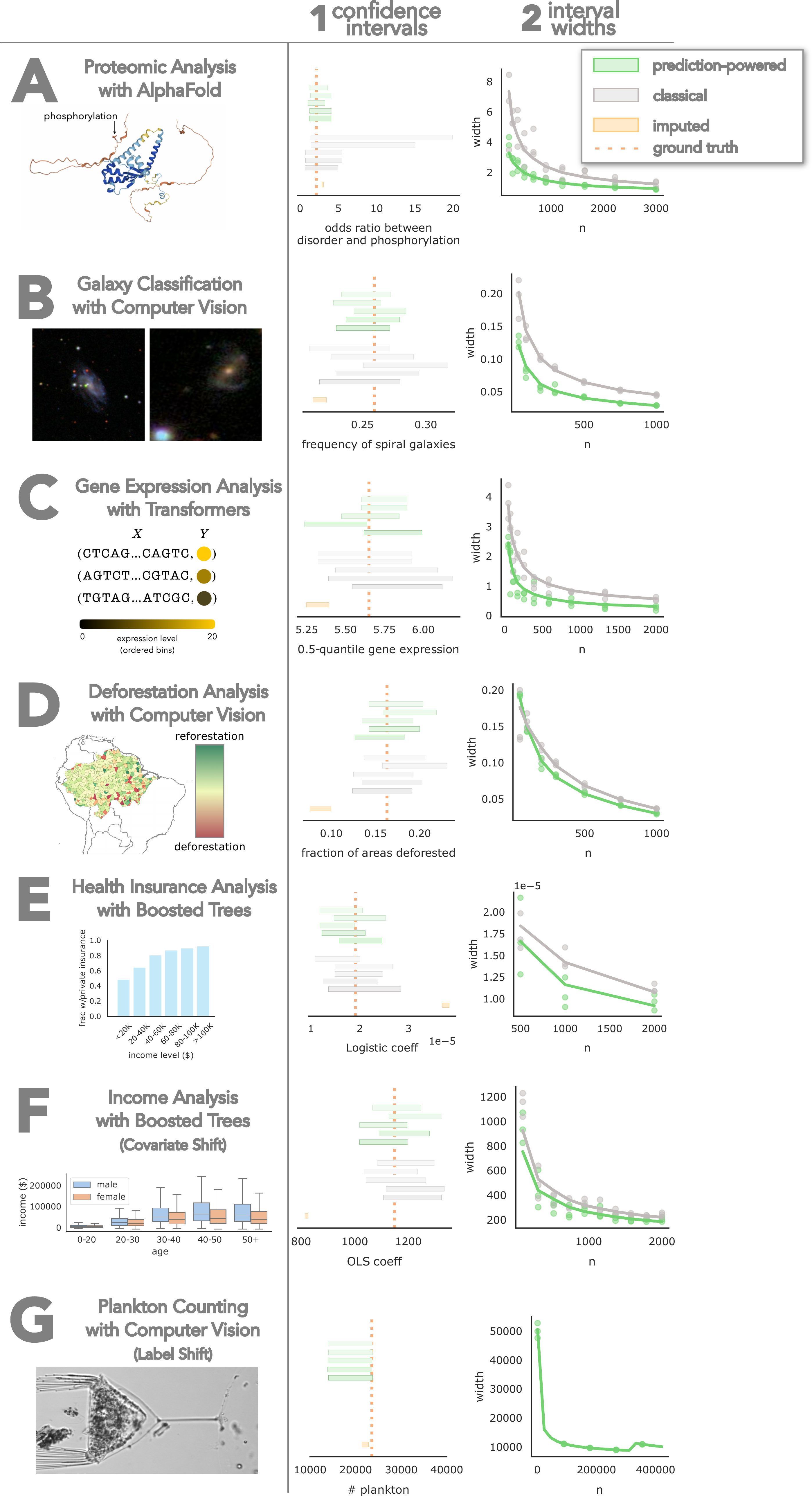}
    \caption{\textbf{Comparison of prediction-powered, classical, and imputation approaches.} Each row (A-G) is a different application. Panel (1) plots five randomly chosen intervals for the prediction-powered and classical approaches, and the imputed interval (which is deterministic). Panel (2) plots the average interval width and the width in the five randomly chosen trials, for varying $n$.}
    \label{fig:big-figure}
\end{figure}

\section{Applications}
\label{sec:experiments}

\begin{table}[t]
\centering
\begin{tabular}{l|cc}
\hline
\textbf{Problem} & \textbf{Prediction-powered} & \textbf{Classical} \\
\hline
\textbf{A} Proteomic analysis with AlphaFold & \( n = 316 \) & \( n = 799 \) \\
\textbf{B} Galaxy classification with computer vision & \( n = 189 \) & \( n = 449 \) \\
\textbf{C} Gene expression analysis with transformers & \( n = 764 \) & \( n = 900 \) \\
\textbf{D} Deforestation analysis with computer vision & \( n = 21 \) & \( n = 35 \) \\
\textbf{E} Health insurance analysis with boosted trees & \( n = 5569 \) & \( n = 6653 \) \\
\textbf{F} Income analysis with boosted trees & \( n = 177 \) & \( n = 282 \) \\
\hline
\end{tabular}
\caption{Number of labeled examples needed to make a discovery with prediction-powered inference and classical inference. The rows (A to F) correspond to the application domains from Figure \ref{fig:big-figure}. For each application, a null hypothesis about $\theta^*$ is tested at level 95\%.}
\label{table:sample-size}

\end{table}

We demonstrate prediction-powered inference on real tasks. In each, we compute a prediction-powered confidence interval for an estimand and compare it to intervals obtained through the classical approach and the imputation approach. In all cases, we show that the imputation approach, which uses machine-learning predictions without accounting for prediction errors, does not contain the true value of the estimand. We compare the widths of the two valid approaches, prediction-powered and classical, as a function of the amount of labeled data used. In addition, we compare the number of labeled examples needed to reject a null hypothesis at level $1-\alpha=95\%$
with high probability. Each trial randomly splits the data into a labeled dataset and an unlabeled dataset. The results are given in Figure \ref{fig:big-figure} and Table \ref{table:sample-size}.

See \cite{ppipy} for a Python package implementing prediction-powered inference, which contains code for reproducing the experiments. Each application below comes with a corresponding Jupyter notebook that can be accessed by clicking these icons: \jupyter{https://github.com/aangelopoulos/ppi_py}.
We packaged the data in such a way that the reader can run the notebooks on their local machine without downloading large datasets.

\subsection{Relating protein structure and post-translational modifications \texorpdfstring{\jupyter{https://github.com/aangelopoulos/ppi_py/blob/main/examples/alphafold.ipynb}}{jupyter}}
\label{sec:protein}

The goal is to characterize whether various types of post-translational modifications (PTMs) occur more frequently in intrinsically disordered regions (IDRs) of proteins \cite{iakoucheva2004importance}. Recently, Bludau et al. \cite{bludau2022structural} studied this relationship on an unprecedented proteome-wide scale by using structures predicted by AlphaFold \cite{jumper2021highly} to predict IDRs, in contrast to previous work which was limited to far fewer experimentally derived structures. 

To quantify the association between PTMs and IDRs, the authors applied the imputation approach: they computed the odds ratio between AlphaFold-based IDR predictions and PTMs on a dataset of hundreds of thousands of protein sequence residues \cite{uniprot2015uniprot}. Using prediction-powered inference, we can combine AlphaFold-based predictions together with gold-standard IDR labels to give a confidence interval for the true odds ratio that is statistically valid, in contrast with the interval constructed with the imputation approach, and smaller than the interval constructed using the classical approach.

We use the fact that the odds ratio, $\theta^*$, between whether or not a protein residue is part of an IDR, $Y \in \{0, 1\}$, and whether or not it has a PTM, $Z \in \{0, 1\}$, can be written as a function of two means:
\begin{align}
\label{eq:odds-ratio}
\theta^* 
= \frac{\mu_1 / (1 - \mu_1)}{\mu_0 / (1 - \mu_0)},
\end{align}
where $\mu_1 = P(Y = 1 \mid Z = 1)$ and $\mu_0 = P(Y = 1 \mid Z = 0)$.
We therefore proceed by constructing $1 - \alpha / 2$ prediction-powered confidence intervals for $\mu_0$ and $\mu_1$, denoted $\CPP_0 = [l_0, u_0]$ and $\CPP_1 = [l_1, u_1]$, respectively.
% , using betting-based concentration inequalities \cite{waudby2020variance}.
We then propagate $\CPP_0$ and $\CPP_1$ through the odds-ratio formula \eqref{eq:odds-ratio} to get the following confidence interval:
\begin{align}
\label{eq:odds-ratio-ci}
    \CPP = \left \{\frac{c_1}{1 - c_1} \cdot \frac{1 - c_0}{c_0}: c_0 \in \CPP_0, c_1 \in \CPP_1 \right \} = \left(\frac{l_1}{1 - l_1} \cdot \frac{1 - u_0}{u_0}, \frac{u_1}{1 - u_1} \cdot \frac{1 - l_0}{l_0} \right).
\end{align}
By a union bound, $\CPP$ contains $\theta^*$ with probability at least $1 - \alpha$.

We have 10803 data points from Bludau et al. \cite{bludau2022structural}. For each of 100 trials, we randomly sample $n$ points to serve as the labeled dataset and treated the remaining 
$N=10803 - n$ points as the unlabeled dataset for which we do not observe the IDR labels. For all values of $n$, the prediction-powered confidence intervals were smaller than classical intervals; see row A in Figure \ref{fig:big-figure}. Often, the classical intervals were large enough that they contained the odds ratio value of one, which means the direction of the association could not be determined from the confidence interval. On the other hand, the imputed confidence interval was far too small and significantly overestimated the true odds ratio. To reject the null hypothesis that the odds ratio is no greater than one, prediction-powered inference required $n=316$ labeled observations, and the classical approach required $n=799$ labeled observations; see row A in Table~\ref{table:sample-size}.

\subsection{Galaxy classification \texorpdfstring{\jupyter{https://github.com/aangelopoulos/ppi_py/blob/main/examples/galaxies.ipynb}}{jupyter}}
\label{sec:galaxy}

The goal is to determine the demographics of galaxies with spiral arms, which are correlated with star formation in the discs of low-redshift galaxies, and therefore, contribute to the understanding of star formation in the Local Universe. A large citizen science initiative called Galaxy Zoo 2 \cite{willett2013galaxy} has collected human annotations of roughly 300000 images of galaxies from the Sloan Digital Sky Survey \cite{york2000sloan} with the goal of measuring these demographics. We seek to explore the use of machine learning to improve the effective sample size and decrease the requisite number of human-annotated galaxies. 

We focus on estimating the fraction of galaxies with spiral arms. We have 1364122 labeled galaxy images from Galaxy Zoo 2, from which we simulate labeled and unlabeled datasets as follows. For each of 100 trials, we randomly sample $n$
points to serve as the labeled dataset and use the remaining 
$N = 1364122 - n$
points as the unlabeled dataset. We then use the algorithm for prediction-powered mean estimation to construct intervals. The prediction-powered confidence intervals for the mean are consistently much smaller than the classical intervals while retaining validity, and the imputation strategy fails to cover; see Figure \ref{fig:big-figure}, row B. To reject the null hypothesis that the fraction of galaxies with spiral arms is at most 0.2, prediction-powered inference requires $n=189$ labeled examples, and classical inference requires $n=449$ examples; see Table~\ref{table:sample-size}, row B.

\subsection{Distribution of gene expression levels \texorpdfstring{\jupyter{https://github.com/aangelopoulos/ppi_py/blob/main/examples/gene_expression.ipynb}}{jupyter}}
\label{sec:gene}

Next, we construct prediction-powered confidence intervals on quantiles that characterize how a population of promoter sequences affects gene expression. Recently, Vaishnav et al. \cite{vaishnav2022evolution} trained a state-of-the-art transformer model to predict the expression level of a particular gene induced by a promoter sequence. They used the model's predictions to study the effects of promoters—for example, by assessing how quantiles of predicted expression levels differ between different populations of promoters. 

Here we focus on estimating different quantiles of gene expression levels induced by native yeast promoters. We have 61150 labeled native yeast promoter sequences from Vaishnav et al. \cite{vaishnav2022evolution}, from which we simulate labeled and unlabeled datasets as follows. For each of 100 trials, we randomly sample $n$ points to serve as the labeled dataset and use the remaining $N = 61150 - n$ points as the unlabeled dataset. We then use the second and third row of Table 1 to construct prediction-powered intervals for the median, as well as the 25\%- and 75\%-quantiles, of the expression levels. The prediction-powered confidence intervals for all three quantiles are much smaller than the classical intervals for all values of $n$. See row C in Figure \ref{fig:big-figure} for the results for the median, and Figure \ref{fig:gene-expression-quantiles} in Appendix \ref{app:genes} for the other two quantiles. We also evaluate the number of labeled examples required by prediction-powered inference and classical inference, respectively, to reject the null hypothesis that the median gene expression level is at most five. Prediction-powered inference requires $n = 764$
examples and classical inference requires $n = 900$ examples; see row C in Table~\ref{table:sample-size}.

\subsection{Estimating deforestation in the Amazon \texorpdfstring{\jupyter{https://github.com/aangelopoulos/ppi_py/blob/main/examples/forest.ipynb}}{jupyter}}
\label{sec:forest}

The goal is to estimate the fraction of the Amazon rainforest lost between 2000 and 2015. Gold-standard deforestation labels for parcels of land are scarce, having been collected largely through field visits, an expensive process impractical for large areas \cite{bullock2020satellite}. However, machine-learning predictions of forest cover based on satellite imagery are readily available for the entire Amazon \cite{sexton2013global}.

We begin with 1596 gold-standard deforestation labels for parcels of land in the Amazon. For each of 100 trials, we randomly sample $n$
data points to serve as the labeled dataset and use the remaining data points as the unlabeled dataset. We use the first row of Table 1 to construct the prediction-powered intervals. The imputation approach yields a small confidence interval that fails to cover the true deforestation fraction. The classical approach does cover the truth at the expense of a wider interval and, accordingly, diminished inferential power. The prediction-powered intervals are smaller than the classical intervals and retain validity; see row D in Figure\ref{fig:big-figure}. We also compare the number of gold-standard deforestation labels required by prediction-powered inference and the classical approach to reject the null hypothesis that there is no deforestation. We obtain 
$n=21$
labels for prediction-powered inference and $n=35$ labels for the classical approach; see row D in Table~\ref{table:sample-size}.

\subsection{Relationship between income and private health insurance \texorpdfstring{\jupyter{https://github.com/aangelopoulos/ppi_py/blob/main/examples/census_healthcare.ipynb}}{jupyter}}
\label{sec:healthcare}

The goal is to investigate the quantitative effect of income on the procurement of private health insurance using US census data. Concretely, we use the Folktables interface \cite{ding2021retiring} to download census data from California in the year 2019 (378817 individuals).

As the labeled dataset with the health insurance indicator, we randomly sample $n$
census entries. The remaining data is used as the unlabeled dataset.  We use a gradient-boosted tree \cite{chen2016xgboost} trained on the previous year's data to predict the health insurance indicator in 2019. We construct a prediction-powered confidence interval on the logistic regression coefficient using the fifth row of Table~\ref{table:grad-rec}. Results in row E in Figure\ref{fig:big-figure} show that prediction-powered inference covers the ground truth, the classical interval is wider, and the imputation strategy fails to cover. We also compare the number of gold-standard labels required by prediction-powered inference and the classical approach to reject the null hypothesis that the logistic regression coefficient is no greater than $1.5 \cdot 10^{-5}$. We observe a significant sample size reduction with prediction-powered inference, which requires $n=5569$
labels, whereas classical inference requires $n= 6653$ labels.

\subsection{Relationship between age and income in a covariate-shifted population \texorpdfstring{\jupyter{https://github.com/aangelopoulos/ppi_py/blob/main/examples/census_income_covshift.ipynb}}{jupyter}}
\label{sec:covshift}

The goal is to investigate the relationship between age and income using US census data. We use the same dataset as in the previous experiment, but the features are age and sex, and the target is yearly income in dollars. Furthermore, we introduce a shift in the distribution of the covariates between the gold-standard and unlabeled datasets by randomly sampling the unlabeled dataset with sampling weights 0.8 for females and 0.2 for males.

We used a gradient-boosted tree \cite{chen2016xgboost} trained on the previous year's raw data to predict the income in 2019. We construct a prediction-powered confidence interval on the ordinary least squares regression coefficient using the covariate-shift-robust version of prediction-powered inference from Corollary \ref{cor:covariate-shift}. Results in row F in Figure\ref{fig:big-figure} show that prediction-powered inference covers the ground truth, the classical interval is wider, and the imputation strategy fails to cover. We also compare the number of gold-standard labels required by prediction-powered inference and the classical approach to reject the null hypothesis that the OLS regression coefficient is no greater than 
800. We observe a significant sample size reduction with prediction-powered inference, which requires 
$n=177$ labels, whereas classical inference requires 
$n=282$ labels. 

\subsection{Counting plankton \texorpdfstring{\jupyter{https://github.com/aangelopoulos/ppi_py/blob/main/examples/plankton.ipynb}}{jupyter}}
\label{sec:plankton}

Assessment of the increases in phytoplankton growth during springtime warming is important for the study of global biogeochemical cycling in response to climate change. We counted the number of plankton observed by the Imaging FlowCytobot \cite{olson2003automated,orenstein2015whoi}, an automated, submersible flow cytometry system, at Woods Hole Oceanographic Institution in the year 2014. We have access to data from 2013, which are labeled, and we impute the 2014 data with machine-learning predictions from a state-of-the-art ResNet fine-tuned on all data up to and including 2012. The 
$X_i$
 are images of organic matter taken by the FlowCytobot and the 
$Y_i$
 are one of \texttt{\{detritus, plankton\}}, where \texttt{detritus} represents unspecified organic matter. 
The labeled dataset consist of 421238 image--label pairs from 2013 and we receive 329832 labeled images from 2014. We use the data from 2014 as our unlabeled data and confirm our results against those that were hand-labeled. The years 2013 and 2014 have a distribution shift, primarily caused by the change in the base frequency of plankton observations with respect to detritus. To apply prediction-powered inference to count the number of plankton recorded in 2014, we use the label-shift-robust technique described in Theorem \ref{thm:label-shift}. The results in row G in Figure \ref{fig:big-figure} show that prediction-powered inference covers the ground truth and the imputation strategy fails to cover.

\section{Extensions}
\label{sec:methods-details}

We demonstrate that the framework of prediction-powered inference is applicable beyond inference under i.i.d. observations and convex losses studied in Section \ref{sec:convex}. First, we provide a strategy for prediction-powered inference when $\theta^*$ can be expressed as the optimum of any optimization problem, not necessarily a convex one. Then, we discuss prediction-powered inference under certain forms of distribution shift. We end with a brief discussion of a natural estimation strategy suggested by prediction-powered inference.

\subsection{Beyond convex estimation}
\label{sec:nonconvex}

The tools developed in Section \ref{sec:convex} were tailored to unconstrained convex optimization problems. In general, however, inferential targets can be defined in terms of nonconvex losses or they may have (possibly even nonconvex) constraints. For such general optimization problems, we cannot expect the condition \eqref{eq:gradient-zero} to hold. In this section we generalize our approach to a broad class of risk minimizers:
\begin{equation}
\label{eq:m-est-target}
\theta^* = \argmin_{\theta\in\Theta} \E [\ell_\theta(X_i,Y_i)],
\end{equation}
where $\ell_\theta:\X\times\Y\rightarrow\R$ is a possibly nonconvex loss function and $\Theta$ is an arbitrary set of admissible parameters. As before, if $\theta^*$ is not a unique minimizer, our method will return a set that contains all minimizers.

The problem~\eqref{eq:m-est-target} subsumes all previously studied settings. Indeed, when the loss $\ell_\theta$ is convex and subdifferentiable and $\Theta = \R^p$ for some $p$---which is the case for all problems previously studied---$\theta^*$ can be equivalently characterized via the condition \eqref{eq:gradient-zero}. In this section we provide a solution that can handle problems of the form \eqref{eq:m-est-target} in full generality. We note, however, that the solution does not reduce to the one in Section \ref{sec:convex} for convex estimation problems, and we expect the method from Section \ref{sec:convex} to be more powerful for convex estimation problems with low-dimensional rectifiers.

To correct the imputation approach, we rely on the following rectifier:
\begin{equation}
\label{eq:rec-m-est}
\rec_{\theta} =   \E\left[ \ell_\theta(X_i,  Y_i) -  \ell_\theta( X_i,  f(X_i)) \right].
\end{equation}
Notice that the rectifier \eqref{eq:rec-m-est} is always one-dimensional, while the rectifier \eqref{eq:convex-rectifier} was $p$-dimensional.

One key difference relative to the approach of Section \ref{sec:convex} is that we have an additional step of data splitting.
We need the additional step because, unlike in convex estimation where we know $\E [g_{\theta^*}(X_i,Y_i)] = 0$, for general problems we do not know the value of $\E [\ell_{\theta^*}(X_i,Y_i)]$. 
To circumvent this issue, we estimate $\E [\ell_{\theta^*}(X_i,Y_i)]$ by approximating $\theta^*$ with an imputed estimate on the first $N/2$ unlabeled data points (for simplicity, take $N$ to be even). To state the main result, we define
$$\thetaf = \argmin_{\theta\in\Theta} \frac{2}{N} \sum_{i=1}^{N/2} \ell_\theta(\tilde X_i,  f(\Xt_i)), \quad \tilde L^f(\theta) := \frac{2}{N} \sum_{i=N/2 + 1}^{N} \ell_\theta(\tilde X_i, f(\Xt_i)).$$

\begin{theorem}[General risk minimization]
\label{thm:m-estimators}
Fix $\alpha\in(0,1)$ and $\delta\in(0,\alpha)$. Suppose that, for any $\theta\in\Theta$, we can construct $\left(\cR_{\delta/2}^l(\theta), \cR_{\delta/2}^u(\theta)\right)$ and $\left(\T_{\frac{\alpha-\delta}{2}}^l(\theta),\T_{\frac{\alpha-\delta}{2}}^u(\theta) \right)$ such that 
\begin{align*}
    &P\left(\rec_{\theta} \leq \cR_{\delta/2}^u(\theta) \right) \geq 1-\delta/2; \quad P\left(\rec_{\theta} \geq \cR_{\delta/2}^l(\theta) \right) \geq 1-\delta/2;\\
    &P\left(\tilde L^f(\theta) -\E[\ell_\theta(X_i,f(X_i))] \leq \T_{\frac{\alpha-\delta}{2}}^u(\theta) \right) \geq 1-\frac{\alpha-\delta}{2}; P\left(\tilde L^f(\theta) -\E[\ell_\theta(X_i,f(X_i))] \geq \T_{\frac{\alpha-\delta}{2}}^l(\theta) \right) \geq 1-\frac{\alpha-\delta}{2}.
\end{align*}
Let
$$\CPP_\alpha = \left\{\theta\in\Theta : \tilde L^f(\theta) \leq   \tilde L^f(\thetaf) - \cR_{\delta/2}^l(\theta) + \cR_{\delta/2}^u(\thetaf) + \T_{\frac{\alpha-\delta}{2}}^u(\theta) - \T_{\frac{\alpha-\delta}{2}}^l(\thetaf)  \right\}.$$
Then, we have
$$P\left( \theta^* \in \CPP_\alpha \right) \geq 1-\alpha.$$
\end{theorem}
\noindent For example, if the loss $\ell_\theta(x,y)$ takes values in $[0,B]$ for all $x,y$, then we can set $\T_{\alpha-\delta}(\theta) = B\sqrt{\frac{\log(1/(\alpha-\delta))}{N}}$. The validity of this choice follows by Hoeffding's inequality.

\paragraph{Mode estimation.}
A commonplace inference task that does not fall under convex estimation is the problem of estimating the mode of the outcome distribution. When the outcome takes values in a discrete set $\Theta$, this can be done by using the loss function $\ell_\theta(y) = \ind{y \neq \theta}, \theta\in\Theta$. A generalization of this approach to continuous outcome distributions is obtained by defining the loss
$\ell_\theta(y) = \ind{|y - \theta| > \eta}$,
for some width parameter $\eta>0$. The target of inference is thus the point $\theta\in\R$ that has the most probability mass in its $\eta$-neighborhood, $\theta^* = \argmin_{\theta\in\R} P\left(|Y_i - \theta| > \eta\right)$.
Theorem~\ref{thm:m-estimators} applies directly in both the discrete and continuous cases.

\paragraph{Tukey's biweight robust mean.} The Tukey biweight loss function is a commonly used loss in robust statistics that results in an outlier-robust mean estimate. It behaves approximately like a quadratic near the origin and is constant far away from the origin. Formally, Tukey's biweight loss function is given by
$$\ell_\theta(y) =
\begin{cases}
\frac{c^2}{6}\left(1 - \left(1 - \frac{\left(y-\theta\right)^2}{c^2}\right)^3 \right), &|y-\theta|\leq c,\\
\frac{c^2}{6}, &\text{otherwise},
\end{cases}$$
where $c$ is a user-specified tuning parameter. It is not hard to see that the function $\ell_\theta(y)$ is nonconvex and hence not amenable to the analysis in Section \ref{sec:convex}; however, Theorem~\ref{thm:m-estimators} applies.

% \paragraph{\clara{Trimmed mean estimation.}}
% The trimmed mean is a common robust measure of distribution location.
% The inference target is $\theta^* = \argmin_{\theta\in\R} \E [l_\theta(\tY_1)]$, where
% $$\ell_\theta(y) =
% \begin{cases}
% \frac{1}{2}(y - \theta)^2, & |y - \theta | \leq c \\
% 0 & \mathrm{otherwise}.
% \end{cases}$$
% for some constant $c \in \mathbb{R}$.

\paragraph{Model selection.} Nonconvex risk minimization problems are ubiquitous in model selection. For example, a common model selection strategy is best subset selection, which optimizes the squared loss, $\ell_\theta(x,y) = (y-x^\top\theta)^2$, subject to the constraint $\Theta = \{\theta\in\R^d:\|\theta\|_0 \leq k\}$.
Here, $\Theta$ is the space of all $k$-sparse vectors for a user-chosen parameter $k$. Even though the loss function is convex, $\Theta$ is a nonconvex constraint set and hence we cannot rely on the condition \eqref{eq:gradient-zero} to find the minimizer.
However, Theorem~\ref{thm:m-estimators} still applies.

\subsection{Inference under distribution shift}
\label{sec:dist-shift}

In Section \ref{sec:convex} we focused on forming prediction-powered confidence intervals when the labeled and unlabeled data come from the same distribution.
Herein, we extend our tools to the case where the labeled data $(X,Y)$ comes from $\P$ and the unlabeled data $(\Xt,\Yt)$---which defines the target of inference $\theta^*$---comes from $\Q$, and these are related by either a label shift or a covariate shift. For covariate shift, we handle all estimation problems previously studied; for label shift, we handle certain types of linear problems.

We will write $\E_{\Q},\E_{\P}$, etc to indicate which distribution the data inside the expectation is sampled from.

\subsubsection{Covariate shift}

First, we assume that $\mathbb{Q}$ is a known \emph{covariate shift} of $\P$. That is, if we denote by $\Q = \Q_X \cdot \Q_{Y \mid X}$ and $\P = \P_X \cdot \P_{Y \mid X}$ the relevant marginal and conditional distributions, we assume that $\Q_{Y \mid X} = \P_{Y \mid X}$. As in previous sections, we consider estimands of the form
\begin{equation}
\label{eq:covariate_shift_estimands}
\theta^* = \argmin_{\theta\in \Theta} \E_{\Q} [\ell_\theta(X_i,Y_i)].
\end{equation}
% for some loss function $\ell_\theta$.

Estimands of the form \eqref{eq:covariate_shift_estimands} can be related to risk minimizers on $\P$ using the Radon-Nikodym derivative. In particular, suppose that $\Q_X$ is dominated by $\P_X$ and assume that the Radon-Nikodym derivative $w(x) = \frac{\Q_X}{\P_X}(x)$ is known. Then, we can rewrite \eqref{eq:covariate_shift_estimands} as 
$$\theta^* = \argmin_{\theta\in \Theta} \E_{\P} [\ell_\theta^w(X_i,Y_i)],$$
where $\ell_\theta^w(x,y) = w(x) \ell_\theta(x,y)$. In words, risk minimizers on $\Q$ can simply be written as risk minimizers on $\P$, but with a reweighted loss function. This permits inference on the rectifier to be based on  data sampled from $\P$ as before. For concreteness, we explain the approach in detail for convex risk minimizers. Let
\begin{equation}
\label{eq:rec-covariate-shift}
\recw_{\theta} = \E_\P \left[g_\theta^w(X_i,Y_i) - g_\theta^w (X_i,f(X_i))\right],
\end{equation}
where $g_\theta^w (x,y) = g_\theta(x,y)\cdot w(x)$ and $g_\theta$ is a subgradient of $\ell_\theta$ as before. A confidence set for the above rectifier suffices for prediction-powered inference on $\theta^*$.

\begin{cor}[Covariate shift]
\label{cor:covariate-shift}
Suppose that the problem \eqref{eq:covariate_shift_estimands} is a nondegenerate convex estimation problem. Fix $\alpha\in(0,1)$ and $\delta\in(0,\alpha)$.
Suppose that, for any $\theta\in\R^p$, we can construct $\cR_\delta(\theta)$ and $\T_{\alpha-\delta}(\theta)$ satisfying
$$P\left(\recw_{\theta} \in \cR_{\delta}(\theta) \right) \geq 1-\delta; \quad P\left(\E [g^w_\theta(X_i,f(X_i))] \in \T_{\alpha-\delta}(\theta) \right) \geq 1-(\alpha-\delta).$$
Let $\CPP_\alpha = \left\{\theta : 0 \in \cR_\delta(\theta) + \T_{\alpha-\delta}(\theta)\right\}$, where $+$ denotes the Minkowski sum. Then,
\begin{equation}
    P( \theta^* \in \CPP_\alpha ) \geq 1-\alpha.
\end{equation}
\end{cor}

\noindent The same reweighting principle can be used to handle nonconvex risk minimizers as in Section~\ref{sec:nonconvex}.

\subsubsection{Label shift}

Next, we analyze classification problems where the proportions of the classes in the labeled data is different from those in the unlabeled data.
This problem has been studied before in the literature on domain adaptation, e.g. by Lipton et al.~\cite{lipton2018detecting}, but our treatment focuses on the formation of confidence intervals.
Formally, let $\Y = \{1, ..., K\}$ be the label space and assume that $\Q_{X|Y} = \P_{X|Y}$. We consider estimands of the form
$$\theta^* = \E_{\Q_Y}[\nu(Y)],$$
where $\nu:\Y\rightarrow \R$ is a fixed function. For example, choosing $\nu(y) = \ind{y = k}$ for some $k\in[K]$ asks for inference on the proportion of instances that belong to class $k$. 

Using an analogous decomposition to the one for mean estimation, we can write
$$\theta^* = \E_{\Q_f} [\nu(f)] + (\E_{\Q_Y}[ \nu(Y)] - \E_{\Q_f}[\nu(f)]) = \thetaflimit + \rec,$$
where $\Q_f$ denotes the distribution of $f(X), X\sim \Q_X$. The quantity $\thetaflimit$ can be estimated using the unlabeled data from $\Q$ and the model.
Estimating the quantity $\rec$ using observations from $\P$ will require leveraging the structure of the distribution shift. Central to our analysis will be the confusion matrix
\begin{equation}
    \K_{j,l} = \Q\left( f(X) = j \; \Big| \; Y = l \right), \; j,l \in [K].
\end{equation}
The label-shift assumption implies that $\K_{j,l} = \P\left( f(X) = j \; | \; Y =l \right)$, which can be estimated from labeled data sampled from $\P$.
In particular, we estimate $\K$ from the labeled data as
\begin{equation}
    \widehat{\K}_{j,l} = \frac{1}{n(l)} \sum_{i=1}^n \ind{ f(X_i) = j, Y_i = l }, \text{ where } n(l) = \sum_{i=1}^n \ind{Y_i = l}.
\end{equation}
Similarly, we can estimate $\Q_f(k),k\in[K]$ as
$$\widehat\Q_f(k) = \frac{1}{N} \sum_{i=1}^N \ind{f(\Xt_i) = k}.$$
Treating $\Q_f$ and $\Q_Y$ as vectors, notice that we can write $\Q_f = \K \Q_Y$, and hence $\Q_Y = \K^{-1}\Q_f$. This leads to a natural estimate of $\Q_Y$, $\widehat \Q_Y = \widehat \K^{-1} \widehat \Q_f$. Below, we use these quantities to construct a prediction-powered confidence interval for $\theta^* = \E_{\Q_Y} [\nu(Y)]$.

\begin{theorem}[Label shift]
\label{thm:label-shift}
Fix $\alpha\in(0,1)$ and $\delta\in(0,\alpha)$. Let
$$\CPP_\alpha = \left(\E_{\widehat \Q_Y} [\nu(Y)] \pm \left(\max_{l,k\in[K]} \max_{p\in C_{l,k}} |\widehat \K_{l,k} - p| + \sqrt{\frac{1}{2N} \log\frac{2}{\alpha-\delta}} \right) \right),$$
where
$$C_{l,k} = \left\{p: n(k) \widehat \K_{l,k} \in \Big[ F^{-1}_{\mathrm{Binom}(n(k),p)}\left(\frac{\delta}{2K^2}\right), F^{-1}_{\mathrm{Binom}(n(k),p)}\left(1-\frac{\delta}{2K^2}\right) \Big] \right\}$$
and $F_{\mathrm{Binom}(n(k),p)}$ denotes the Binomial CDF.
Then,
\begin{equation*}
P(\theta^* \in \CPP_\alpha)\geq 1-\alpha.
\end{equation*}
\end{theorem}
\noindent Naturally, the confidence interval becomes more conservative as the number of classes grows. Also, the power of the bound depends on the smallest number of instances observed for a particular class.

\subsection{Prediction-powered point estimate}
\label{subsec:point-estimate}

Prediction-powered inference suggests a natural approach to constructing point estimates as well.
Define the \emph{rectified} loss function as
\begin{equation}
    \label{eq:rectified-loss}
    L^{\rm PP}(\theta) = \frac{1}{N} \sum\limits_{i=1}^N \ell_{\theta}(\tX_i, f(\tX_i)) + \frac{1}{n} \sum\limits_{i=1}^n \left( \ell_{\theta}(X_i,Y_i) - \ell_{\theta}(X_i, f(X_i)) \right).
\end{equation}
The expected value of the rectified loss is equal to the true population loss that $\theta^*$ minimizes: $\E[L^{\rm PP}(\theta)] = \E[\ell_\theta(X_i,Y_i)]$.
We define the prediction-powered point estimate as the minimizer of the rectified loss:
\begin{equation}
    \hat\theta^{\rm PP} = \arg\min_{\theta} L^{\rm PP}(\theta).
\end{equation}
% Differentiating~\eqref{eq:rectified-loss} yields $\nabla L^{\rm PP}(\theta) = \hat{g}_{\theta}^f + \rechat_{\theta}$.
The confidence intervals formed in Algorithms~\ref{alg:means}-~\ref{alg:general-convex} were implicitly based on the gradient of the rectified loss, $\nabla L^{\rm PP}(\theta) = \hat{g}_{\theta}^f + \rechat_{\theta}$. More precisely, they all used $\nabla L^{\rm PP}(\theta)$ as a statistic for testing whether $\E[\nabla \ell_\theta(X_i,Y_i)] =~0$. Notice that the prediction-powered point estimate is always contained in the constructed confidence intervals, since it satisfies $\hat{g}_{\theta}^f + \rechat_{\theta} = 0$.

\section{Acknowledgments}
We would like to thank Amit Kohli for suggesting the term ``rectifier,'' Eric Orenstein for helpful discussions related to the WHO-I Plankton dataset, Philip Stark and the San Francisco Department of Elections for helping us find the ballot dataset, and Sherrie Wang for help with the remote sensing experiment.
This work was supported in part by the Mathematical Data Science program of the
Office of Naval Research under grant number N00014-21-1-2840 and the National Science Foundation Graduate Research Fellowship.
% \newpage
\printbibliography

\clearpage
\appendix

\section{Prediction-powered p-values}

By relying on the standard duality between confidence intervals and p-values, we can immediately repurpose the presented theory to compute valid prediction-powered p-values.

To formalize this, suppose that we want to test the hull hypothesis $H_0: \theta^*\in\Theta_0$, for some set $\Theta_0 \in \R^p$ (for example, a common choice when $p=1$ is $\Theta_0 = \R_{\leq 0}$). Let $\C_\alpha$ be a valid confidence interval. Then, we can construct a valid p-value as
\begin{equation*}
P = \inf\{\alpha: \theta_0 \not\in \C_\alpha, \forall \theta_0\in\Theta_0\}.
\end{equation*}
A p-value $P$ is valid if it is super-uniform under the null, meaning $P(P \leq u)\leq u$ for all $u\in[0,1]$. This is indeed the case for the p-value defined above, because when $\theta^*\in\Theta_0$, we have
$$P(P\leq u) \leq P(\theta^* \not\in \C_u) \leq u.$$
The first inequality follows by the definition of $P$ and the fact that $\theta^*\in\Theta_0$, and the second inequality follows by the validity of $\C_u$ at level $1-u$. We are implicitly using the fact that $\C_u \subseteq \C_{u'}$ when $u\geq u'$.

The above derivation is a general recipe for deriving p-values from confidence intervals. For the prediction-powered confidence intervals stated in Algorithms \ref{alg:means}-\ref{alg:general-convex} (derived from Theorem \ref{thm:convex-validity_asymptotic}), the corresponding prediction-powered p-value is given by:
$$P^{\rm PP} = \inf\left\{\alpha: |\hat g^f_{\theta_0} + \rechat_{\theta_0}| > w_{\alpha}(\theta_0), ~\forall \theta_0\in\Theta_0 \right\}.$$

Below we state analogues of Algorithms \ref{alg:means}-\ref{alg:ols} when the goal is to compute a prediction-powered p-value.

\begin{algorithm}[H]
\caption{Prediction-powered p-value for the mean}
\label{alg:means_pval}
\begin{algorithmic}[1]
\Require labeled data $(\Xa, \Ya)$, unlabeled features $\tX$, predictor $f$, null set $\Theta_0$
\State $\thetaPP \leftarrow \thetaf - \rechat := \frac{1}{N}\sum_{i=1}^N f(\Xt_i) - \frac{1}{n}\sum_{i=1}^n (f(X_i) - Y_i)$
\State $\hat \sigma_{f}^2 \leftarrow \frac 1 N \sum_{i=1}^N (f(\Xt_i) - \thetaf)^2$
\State $\hat \sigma_{f-Y}^2 \leftarrow \frac 1 n \sum_{i=1}^n (f(X_i) - Y_i - \rechat )^2$
\State Define $w_\alpha := z_{1-\alpha/2} \sqrt{\frac{\hat\sigma_{f-Y}^2}{n} + \frac{\hat\sigma_{f}^2}{N}}$
\Ensure 
prediction-powered p-value $P^{\rm PP} = \inf\{\alpha: \theta_0\not\in(\thetaPP \pm w_\alpha), \forall \theta_0\in\Theta_0\}$
\end{algorithmic}
\end{algorithm}

\vspace{-.3cm}

\begin{algorithm}[H]
\caption{Prediction-powered p-value for the quantile}
\label{alg:quantile_pval}
\begin{algorithmic}[1]
\Require labeled data $(\Xa, \Ya)$, unlabeled features $X'$, predictor $f$, quantile $q$, null set $\Theta_0$
\For{$\theta \in \Theta_0$}
\State $\rechat_\theta \leftarrow \frac{1}{n}\sum_{i=1}^n \left(\ind{Y_i \leq \theta} - \ind{f(X_i) \leq \theta}\right)$
\State $\hat F_\theta  \leftarrow \frac{1}{N}\sum_{i=1}^N  \ind{f(\Xt_i) \leq \theta}$
\State $\hat{\sigma}^2_{\Delta}(\theta)  \leftarrow \frac{1}{n}\sum_{i=1}^n\left(\ind{Y_i \leq \theta} - \ind{f(X_i) \leq \theta} - \rechat_\theta\right)^2$
\State $\hat{\sigma}_{g}^2(\theta) \gets \frac{1}{N}\sum_{i=1}^N\left(\ind{f(\Xt_i) \leq \theta} - \hat F_\theta \right)^2$
\State Define $w_\alpha(
\theta) := z_{1-\alpha/2}\sqrt{\frac{\hat{\sigma}^2_{\Delta}(\theta)}{n} + \frac{\hat{\sigma}^2_{g}(\theta)}{N}}$
\EndFor
\Ensure 
prediction-powered p-value $P^{\rm PP} = \inf\left\{ \alpha:  |\hat F_{\theta_0} + \rechat_{\theta_0} - q|  > w_{\alpha}(\theta_0), \forall \theta_0\in\Theta_0 \right\}$
\end{algorithmic}
\end{algorithm}

\vspace{-.3cm}

\begin{algorithm}[H]
\caption{Prediction-powered p-value for logistic regression coefficients}
\label{alg:logistic_pval}
\begin{algorithmic}[1]
\Require labeled data $(\Xa, \Ya)$, unlabeled features $\Xt$, predictor $f$, null set $\Theta_0$
\State $\rechat_j \leftarrow \frac{1}{n}\sum_{i=1}^n X_{i,j}(f(X_i) - Y_i), \quad j\in[d]$ 
\State $\hat \sigma_{\Delta,j}^2 \leftarrow \frac{1}{n}\sum_{i=1}^n\left(X_{i,j}(f(X_i) - Y_i) - \rechat_j\right)^2, \quad j\in[d]$ 
\For{$\theta \in \Theta_0$}
\State $\hat g^f_{\theta,j} \leftarrow \frac{1}{N}\sum_{i=1}^N \Xt_{i,j}\left(\mu_\theta(\Xt_i) - f(\Xt_i) \right), \quad j\in [d]$, \quad where $\mu_\theta(x) = \frac{1}{1+\exp(-x^\top \theta)}$
\State $\hat \sigma_{g,j}^2(\theta) \leftarrow \frac{1}{N}\sum_{i=1}^N\left(\Xt_{i,j}(\mu_\theta(\Xt_i) - f(\Xt_i)) - \hat g^f_{\theta,j}\right)^2, \quad j\in[d]$ 
\State Define $w_{\alpha,j}(\theta) := z_{1-\alpha/(2d)} \sqrt{ \frac{\hat\sigma_{\Delta,j}^2}{n} +  \frac{\hat \sigma_{g,j}^2(\theta)}{N}}, \quad j\in[d]$
\EndFor
\Ensure 
prediction-powered p-value $P^{\rm PP} = \inf\left\{\alpha:  |g^f_{\theta_0,j} + \rechat_j| > w_{\alpha,j}(\theta_0), \forall j\in[d], \theta_0 \in \Theta_0 \right\}$
\end{algorithmic}
\end{algorithm}

\vspace{-.3cm}

\begin{algorithm}[H]
\caption{Prediction-powered p-value for linear regression coefficients}
\label{alg:ols_pval}
\begin{algorithmic}[1]
\Require labeled data $(X, Y)$, unlabeled features $\Xt$, predictor $f$, coefficient $j^*$, null set $\Theta_0$
\State $\thetaPP \leftarrow \thetaf - \rechat := \Xt^\dagger f(\Xt) - X^\dagger (f(X) - Y)$
\State $\tilde \Sigma \leftarrow \frac 1 N \Xt^\top \Xt$, $\tilde M \leftarrow \frac 1 N \sum_{i=1}^N (f(\Xt_i) - \Xt_i^\top  \thetaf)^2 \Xt_i \Xt_i^\top$
\State $\tilde V \leftarrow (\tilde \Sigma)^{-1} \tilde M (\tilde \Sigma)^{-1}$
\State $\Sigma \leftarrow \frac 1 n X^\top X$,  $M\leftarrow \frac 1 n \sum_{i=1}^n (f(X_i) - Y_i - X_i^\top  \rechat)^2 X_i X_i^\top$
\State $V \leftarrow \Sigma^{-1} M  \Sigma^{-1}$
\State Define $w_{\alpha} := z_{1-\alpha/2} \sqrt{ \frac{ V_{j^*j^*}}{n} + \frac{ \tilde V_{j^*j^*}}{N}}$
\Ensure 
prediction-powered confidence set $\CPP_\alpha = \inf\{\alpha: \theta_0\not\in (\thetaPP_{j^*} \pm w_{\alpha}),\forall \theta_0\in\Theta_0 \}$
\end{algorithmic}
\end{algorithm}

\begin{cor}[Mean p-value]
\label{cor:mean_est_validity_pval} 
Let $\theta^*$ be the mean outcome:
$$\theta^* = \E[Y_i].$$
Then, the prediction-powered p-value in Algorithm \ref{alg:means_pval} is valid: under the null, $\liminf_{n,N\rightarrow\infty}P\left(P^{\mathrm{PP}}\leq u\right) \leq~u,\forall u\in~[0,1]$.
\end{cor}

\begin{cor}[Quantile p-value]
\label{cor:quantile_pval}
Let $\theta^*$ be the $q$-quantile: 
$$\theta^* = \min\{\theta: P(Y_i \leq \theta) \geq q\}.$$
Then, the prediction-powered p-value in Algorithm \ref{alg:quantile_pval} is valid: under the null, $\liminf_{n,N\rightarrow\infty}P\left(P^{\mathrm{PP}}\leq u\right) \leq~u,\forall u\in~[0,1]$.
\end{cor}

\begin{cor}[Logistic regression p-value]
\label{cor:logistic_pval}
Let $\theta^*$ be the logistic regression solution:
$$\theta^* = \argmin_{\theta\in\R^d} \E\left[-Y_i \theta^\top X_i + \log(1 + \exp(\theta^\top X_i))\right].$$
Then, the prediction-powered p-value in Algorithm \ref{alg:logistic_pval} is valid: under the null, $\liminf_{n,N\rightarrow\infty}P\left(P^{\mathrm{PP}}\leq u\right) \leq~u,\forall u\in~[0,1]$.
\end{cor}

\begin{cor}[Linear regression p-value]
\label{cor:ols_pval}
Fix $j^*\in[d]$. Let $\theta^*$ be the linear regression solution:
$$\theta^* = \argmin_{\theta\in\R^d} \E[(Y_i - X_i^\top\theta)^2].$$
Then, 
the prediction-powered p-value in Algorithm \ref{alg:ols_pval} is valid: under the null, $\liminf_{n,N\rightarrow\infty}P\left(P^{\mathrm{PP}}\leq u\right) \leq~u,\forall u\in~[0,1]$.
\end{cor}

\section{Inference on a finite population}
\label{app:finite-pop}

The techniques developed in this paper directly translate to the \emph{finite-population} setting. Here, we treat $(\tilde X,\tilde Y)$ as a fixed finite population consisting of $N$ feature-outcome pairs, without imposing any distributional assumptions on the data points. Analogously to the i.i.d. setting, we observe all features $\tilde X$ and a small set of outcomes. Specifically, we assume that we observe $(\Yt_i)_{i\in\Ical}$, where $\Ical = \{i_1,\dots,i_n\}$ is a uniformly sampled subset of $[N]$ of size $n \ll N$. In this section we adapt all our main results to the finite-population context.

Given a loss function $\ell_\theta$ and parameter space $\Theta$, the target estimand is the risk minimizer we would compute if we could observe the whole population:
\begin{equation}
\label{eq:finite-pop-target}
\theta^* = \argmin_{\theta\in\Theta} \frac 1 N \sum_{i=1}^N \ell_\theta(\Xt_i,\Yt_i).
\end{equation}
The following two subsections mirror the results for convex and nonconvex estimation from the main body of the paper. All results in this section are proved essentially identically as their i.i.d. counterparts.

In what follows, we construct prediction-powered confidence sets $\CPP_\alpha$ assuming a valid confidence set around the rectifier (defined below for the finite-population context). The confidence set for the rectifier can be constructed from $(\Xt_i,\Yt_i)_{i\in\Ical}$ via a direct application of off-the-shelf results outlined in Appendix \ref{app:cis}. In particular, in Proposition \ref{prop:finite-partial-sum-concentration} we state an asymptotically valid interval for the mean based on a finite-population version of the central limit theorem, and in Proposition \ref{prop:wsr-finite-pop} we state a nonasymptotically valid interval for the mean for finite populations due to Waudby-Smith and Ramdas~\cite{waudby2020variance}. The only assumption required to apply the latter is that $g_\theta(\Xt_i,\Yt_i) - g_\theta(\Xt_i,f(\Xt_i))$ has a known bound valid for all $i\in[N]$.

\subsection{Convex estimation}

In the finite-population setting, the mild nondegeneracy condition ensured by convexity takes the form
\begin{equation}
    \label{eq:gradient-zero-finite-pop}
    \frac{1}{N} \sum_{i=1}^N g_{\theta^*}(\tX_i, \tY_i)  = 0,
\end{equation}
where $g_{\theta}$ is a subgradient of $\ell_{\theta}$. The rectifier is thus:
\begin{equation}
\label{eq:convex-rectifier-finite-pop}
    \rec_{\theta} = \frac{1}{N} \sum_{i=1}^N \left(g_\theta(\tX_i, \Yt_i) - g_\theta(\tX_i, f(\Xt_i))\right).
\end{equation}

\begin{theorem}[Convex estimation, finite population]
\label{thm:convex-finite-pop}
Suppose that the convex estimation problem is nondegenerate \eqref{eq:gradient-zero-finite-pop}. Fix $\alpha\in(0,1)$.
Suppose that, for any $\theta\in\R^p$, we can construct $\cR_\alpha(\theta)$ satisfying
$$P\left(\rec_{\theta} \in \cR_{\alpha}(\theta) \right) \geq 1-\alpha.$$
% where $\rec_{\theta}$ is defined in \eqref{eq:convex-rectifier-finite-pop}.
Let $\CPP_\alpha = \left\{\theta : -\frac{1}{N}\sum_{i=1}^N g_\theta(\Xt_i,f(\Xt_i)) \in \cR_\alpha(\theta) \right\}$. Then,
\begin{equation}
    P( \theta^* \in \CPP_\alpha ) \geq 1-\alpha.
\end{equation}
\end{theorem}

We apply Theorem \ref{thm:convex-finite-pop} in the context of mean estimation, quantile estimation, logistic regression, and linear regression. The target estimand $\theta^*$ is defined as in \eqref{eq:finite-pop-target} with the loss function chosen appropriately, as discussed in Section \ref{subsec:explicit_algs}. We remark that, just like in the i.i.d. case, the analysis for linear regression follows a more refined approach, as in the proof of Proposition \ref{prop:ols}.

\begin{cor}[Mean estimation, finite population]
\label{cor:mean-cor-finite-pop}
    Let $\theta^*$ be the mean outcome. Fix $\alpha\in(0,1)$. Suppose that, for any $\theta\in\R$, we can construct an interval $(\cR_\alpha^l, \cR_\alpha^u)$ such that
    $P\left(\rec \in (\cR_\alpha^l, \cR_\alpha^u) \right)\geq 1-\alpha$,
    where
    $$\rec = \frac 1 N \sum_{i=1}^N \left(f(\Xt_i) -\Yt_i \right).$$
    Let
    $$\CPP_\alpha = \left(\frac{1}{N} \sum_{i=1}^N f(\Xt_i) - \cR_\alpha^u, \frac{1}{N} \sum_{i=1}^N f(\Xt_i)  - \cR_\alpha^l \right).$$
    Then,
    $$P\left(\theta^* \in \CPP_{\alpha}\right) \geq 1-\alpha.$$
\end{cor}

\begin{cor}[Quantile estimation, finite population]
\label{cor:quantile_cor-finite-pop}
    Let $\theta^*$ be the $q$-quantile. Fix $\alpha\in(0,1)$. Suppose that, for any $\theta\in\R$, we can construct an interval $(\cR_\alpha^l(\theta), \cR_\alpha^u(\theta))$ such that
    $P\left(\rec_{\theta} \in (\cR_\alpha^l(\theta), \cR_\alpha^u(\theta)) \right)\geq 1-\alpha$,
    where
    $$\rec_{\theta} = \frac 1 N \sum_{i=1}^N \left(\ind{\Yt_i \leq \theta} - \ind{f(\Xt_i) \leq \theta}\right).$$
    Let
    $$\CPP_\alpha = \left\{\theta\in\R:\frac{1}{N} \sum_{i=1}^N \ind{f(\Xt_i) \leq \theta}\in \left(q - \cR_\alpha^u(\theta), q - \cR_\alpha^l(\theta) \right) \right\}.$$
    Then,
    $$P\left(\theta^* \in \CPP_{\alpha}\right) \geq 1-\alpha.$$
\end{cor}

\begin{cor}[Logistic regression, finite population]
\label{cor:logistic-valid-finite-pop}
Let $\theta^*$ be the logistic regression solution. Fix $\alpha\in(0,1)$. Suppose that we can construct $\cR_\alpha^l, \cR_\alpha^u\in\R^d$ such that $P(\rec_j \in (\cR_{\alpha,j}^l, \cR_{\alpha,j}^u), \forall j\in[d]) \geq 1-\alpha$, where
$$\rec = 
\frac 1 N \sum_{i=1}^N \Xt_{i} (f(\Xt_i) - \Yt_i).$$
Let
$$\CPP_\alpha = \left\{\theta \in \R^d: \frac{1}{N}\sum_{i=1}^N \Xt_{i,j} \left(f(\Xt_i) - \frac{1}{1+\exp(-\Xt_i^\top \theta)}  \right) \in \left(\cR_{\alpha,j}^l , \cR_{\alpha,j}^u \right), \forall j\in[d] \right\}.$$
Then,
\begin{equation}
    P( \theta^* \in \CPP_\alpha ) \geq 1-\alpha.
\end{equation}
\end{cor}

\begin{cor}[Linear regression, finite population]
\label{cor:ols-valid-finite-pop}
Let $\theta^*$ be the linear regression solution. Fix $\alpha\in(0,1)$. Suppose that we can construct $\cR_\alpha^l, \cR_\alpha^u\in\R^d$ such that $P(\rec_j \in (\cR_{\alpha,j}^l, \cR_{\alpha,j}^u), \forall j \in[d]) \geq 1-\alpha$, where
$$\rec = \Xt^\dagger (f(\Xt)-\Yt).$$
Let
$$\CPP_\alpha = \left(\Xt^\dagger f(\Xt) - \cR_{\alpha}^u, \Xt^\dagger f(\Xt)  - \cR_{\alpha}^l\right).$$
Then,
\begin{equation}
    P( \theta^* \in \CPP_\alpha ) \geq 1-\alpha.
\end{equation}
\end{cor}

\subsection{Beyond convex estimation}

We now consider general risk minimizers in the finite-population context. The rectifier is equal to:
\begin{equation}
\rec_{\theta} =   \frac{1}{N}\sum_{i=1}^N \left( \ell_\theta(\tilde X_i, \tilde Y_i) -  \ell_\theta(\tilde X_i, f(\Xt_i)) \right).
\end{equation}

Unlike in the i.i.d. setting, there is no need for data splitting because the imputed estimate is deterministic. We let:
$$\tilde L^f(\theta) := \frac{1}{N} \sum_{i=1}^{N} \ell_\theta(\tilde X_i,  f(\Xt_i)); \quad \thetaf = \argmin_{\theta\in\Theta} \tilde L^f(\theta).$$

\begin{theorem}[General risk minimization, finite population]
\label{thm:m-estimators-finite-pop}
Fix $\alpha\in(0,1)$. Suppose that, for any $\theta\in\Theta$, we can construct $(\cR_{\alpha/2}^l(\theta), \cR_{\alpha/2}^u(\theta))$ such that 
\begin{align*}
    &P\left(\rec_{\theta} \leq \cR_{\alpha/2}^u(\theta) \right) \geq 1-\alpha/2; \quad P\left(\rec_{\theta} \geq \cR_{\alpha/2}^l(\theta) \right) \geq 1-\alpha/2.
\end{align*}
Let
$$\CPP_\alpha = \left\{\theta\in\Theta : \tilde L^f(\theta) \leq  \tilde L^f(\thetaf) - \cR_{\alpha/2}^l(\theta) + \cR_{\alpha/2}^u(\thetaf)  \right\}.$$
Then, we have
$$P\left( \theta^* \in \CPP_\alpha \right) \geq 1-\alpha.$$
\end{theorem}

\section{Deferred theoretical details}

We state an asymptotic counterpart of Theorem \ref{thm:convex-validity} that is used to prove the propositions in Section \ref{subsec:explicit_algs}. Then, we provide nonasymptotically-valid counterparts of the algorithms in Section \ref{subsec:explicit_algs}. Finally, we state the regularity conditions necessary for the guarantees presented in Section \ref{subsec:explicit_algs}.

\subsection{Asymptotic counterpart of Theorem \ref{thm:convex-validity}}
\label{app:asymp-convex-validity}

The following is an asymptotic counterpart of Theorem \ref{thm:convex-validity} that uses the central limit theorem in the confidence set construction. We note the error budget splitting used in Theorem \ref{thm:convex-validity} is in fact not necessary, but we believe that it facilitates exposition when presenting nonasymptotic guarantees. The asymptotic result below is stated without the splitting of the error budget. The proof is stated in Appendix \ref{app:proofs}.

\begin{theorem}[Convex estimation: asymptotic version]
\label{thm:convex-validity_asymptotic}
Suppose that the convex estimation problem is nondegenerate as in \eqref{eq:gradient-zero} and that $\frac{n}{N}\rightarrow p$, for some $p\in(0,1)$. Fix $\alpha\in(0,1)$. For all $\theta\in\R^p$, define
$$\rechat_{\theta} = \frac{1}{n}\sum_{i=1}^n \left(g_\theta(X_i,Y_i) - g_\theta(X_i,f(X_i)) \right); \quad \hat g^f_\theta = \frac{1}{N} \sum_{i=1}^N g_\theta(\Xt_i,f(\Xt_i)).$$
Further, denoting by $g_{\theta,j}(x,y)$ the $j$-th coordinate of $g_\theta(x,y)$, let 
$$\hat\sigma^2_{\Delta,j}(\theta) = \frac{1}{n} \sum_{i=1}^n \left(g_{\theta,j}(X_i,Y_i) - g_{\theta,j}(X_i,f(X_i)) - \rechat_{\theta, j} \right)^2; \quad \hat \sigma^2_{g,j}(\theta) = \frac{1}{N} \sum_{i=1}^N \left(g_{\theta,j}(\Xt_i,f(\Xt_i)) - \hat g^f_{\theta, j}\right)^2,$$
for all $j\in[p]$.
Let $w_{\alpha,j}(\theta) = z_{1-\alpha/(2p)}\sqrt{\frac{\hat\sigma^2_{\Delta,j}(\theta)}{n} + \frac{\hat \sigma^2_{g,j}(\theta)}{N}}$ and $\CPP_\alpha = \left\{\theta : |\rechat_{\theta, j} + \hat g^f_{\theta, j}| \leq w_{\alpha,j}(\theta),~\forall j\in[p]\right\}$. Then,
\begin{equation}
    \liminf_{n,N\rightarrow\infty} P( \theta^* \in \CPP_\alpha ) \geq 1-\alpha.
\end{equation}
\end{theorem}

\subsection{Algorithms with nonasymptotic validity}
\label{app:nonasymptotic-algos}

We state nonasymptotically-valid algorithms for prediction-powered mean estimation, quantile estimation, and logistic regression. Like the methods in Section~\ref{subsec:explicit_algs}, the algorithms rely on the abstract recipe from Theorem~\ref{thm:convex-validity}. The proofs of validity are included in Appendix \ref{app:proofs}.

The following algorithms rely on any off-the-shelf method for computing confidence intervals for the mean. We choose a variance-adaptive confidence interval for the mean due to Waudby-Smith and Ramdas~\cite{waudby2020variance}, which we state in
Algorithm \ref{alg:meanCI}.
We opt to present this construction as the default nonasymptotic confidence interval for the mean because of its strong practical performance.
The only assumption required to apply Algorithm \ref{alg:meanCI} is that the observations are almost surely bounded within a known interval.

\begin{algorithm}[H]
\caption{Prediction-powered mean estimation (nonasymptotic)}
\label{alg:means_nonasymp}
\begin{algorithmic}[1]
\Require labeled data $(\Xa, \Ya)$, unlabeled features $\Xt$, predictor $f$, error levels $\alpha,\delta\in(0,1)$, bound $B$
\State $( f^{l}_{\alpha-\delta},  f^{u}_{\alpha-\delta}) \leftarrow \texttt{MeanCI}\Big(\{f(\Xt_i)\}_{i=1}^N, \mathrm{err} = \alpha - \delta, \mathrm{range} = [0,B]\Big)$
\State $(\cR_\delta^l, \cR_\delta^u) \leftarrow \texttt{MeanCI}\Big(\{f(X_i) - Y_i\}_{i=1}^n, \mathrm{err} = \delta, \mathrm{range} = [-B,B] \Big)$
\Ensure 
prediction-powered confidence set $\CPP_\alpha = \left(f^{l}_{\alpha-\delta} - \cR_\delta^u, f^{u}_{\alpha-\delta} - \cR_\delta^l \right)$
\end{algorithmic}
\end{algorithm}

\begin{cor}[Mean estimation]
\label{cor:means_cor}
    Let $\theta^*$ be the mean outcome~\eqref{eq:mean-outcome}. Suppose that $Y_i,f(X_i)\in[0,B]$ almost surely. Then, the prediction-powered confidence set in Algorithm \ref{alg:means_nonasymp} has valid coverage: $P(\theta^* \in \CPP_\alpha)\geq 1-\alpha$.
\end{cor}

\begin{algorithm}[H]
\caption{Prediction-powered quantile estimation (nonasymptotic)}
\label{alg:quantile_nonasymp}
\begin{algorithmic}[1]
\Require labeled data $(\Xa, \Ya)$, unlabeled features $\Xt$, predictor $f$, quantile $q\in(0,1)$, error levels $\alpha,\delta\in(0,1)$
\State Construct fine grid $\Theta_{\rm grid}$ between $\min_{i\in[N]} f(\Xt_i)$ and $\max_{i\in[N]} f(\Xt_i)$ 
\For{$\theta \in \Theta_{\rm grid}$}
\State $(\cR_\delta^l(\theta), \cR_\delta^u(\theta)) \leftarrow \texttt{MeanCI}\Big(\left\{\ind{Y_i \leq \theta} - \ind{f(X_i) \leq \theta}\right\}_{i=1}^n, \mathrm{err} = \delta, \mathrm{range} = [-1,1]\Big)$
\State $(\hat F^l_{\alpha-\delta}(\theta), \hat F^u_{\alpha-\delta}(\theta))  \leftarrow \texttt{MeanCI} \Big(\left\{\ind{f(\Xt_i) \leq \theta} \right\}_{i=1}^N, \mathrm{err} = \alpha-\delta, \mathrm{range} = [0,1]\Big)$
\EndFor
\Ensure 
prediction-powered confidence set $\CPP_\alpha = \left\{ \theta \in \Theta_{\rm grid} : q \in \left(\hat F^l_{\alpha-\delta}(\theta) + \cR_\delta^l(\theta), \hat F^u_{\alpha-\delta}(\theta) + \cR_\delta^u(\theta) \right) \right\}$
\end{algorithmic}
\end{algorithm}

\begin{cor}[Quantile estimation]
\label{cor:quantile_cor}
    Let $\theta^*$ be the $q$-quantile~\eqref{eq:q-quantile}. Then, the prediction-powered confidence set in Algorithm \ref{alg:quantile_nonasymp} has valid coverage: $P(\theta^* \in \CPP_\alpha)\geq 1-\alpha$.
\end{cor}

\begin{algorithm}[H]
\caption{Prediction-powered logistic regression (nonasymptotic)}
\label{alg:logistic_nonasymp}
\begin{algorithmic}[1]
\Require labeled data $(\Xa, \Ya)$, unlabeled features $\Xt$, predictor $f$, error levels $\alpha,\delta \in(0,1)$, bound $\mathbf{B} = (B_j)_{j=1}^d$ 
\State Construct fine grid $\Theta_{\rm grid} \subset \mathbb{R}^d$ of possible coefficients
\State $(\cR_{\delta,j}^l, \cR_{\delta,j}^u) \leftarrow \texttt{MeanCI}\left( \{X_{i,j}(f(X_i) - Y_i)\}_{i=1}^n, \mathrm{err} = \delta, \mathrm{range} = [-B_j,B_j] \right), \quad j\in[d]$ 
\For{$\theta \in \Theta_{\rm grid}$}
\State $( g_{\alpha-\delta,j}^l(\theta), g_{\alpha-\delta,j}^u(\theta) ) \leftarrow \texttt{MeanCI}\left(\{\Xt_{i,j}\left(\mu_\theta(\Xt_i) - f(\Xt_i)\right)\}_{i=1}^N, \mathrm{err} = \frac{\alpha - \delta}{d}, \mathrm{range} = [-B_j,B_j] \right), j\in~[d]$,\\
\quad \text{ where } $\mu_\theta(x) = \frac{1}{1+\exp(-x^\top \theta)}$
\EndFor
\Ensure 
prediction-powered confidence set $\CPP_\alpha = \left\{\theta\in \Theta_{\rm grid}:  0 \in \left[g_{\alpha-\delta,j}^l(\theta) + \cR_{\delta,j}^l, g_{\alpha-\delta,j}^u(\theta) + \cR_{\delta,j}^u \right], \forall j\in[d] \right\}$
\end{algorithmic}
\end{algorithm}

\begin{cor}[Logistic regression]
\label{cor:logistic_valid}
Let $\theta^*$ be the logistic regression solution \eqref{eq:logistic-sol}. Suppose that $|X_{1,j}|\leq B_j$ and $Y_i,f(X_i)\in[0,1]$ almost surely. Then, the prediction-powered confidence set in Algorithm \ref{alg:logistic_nonasymp} has valid coverage: $P(\theta^* \in \CPP_\alpha)\geq 1-\alpha$.
\end{cor}

We note that there exists an analogous nonasymptotic algorithm for linear regression, however we do not recommend it in practice. The reason is that the refined (but asymptotic) analysis used to prove Proposition~\ref{prop:ols} shows that it is sufficient to analyze a one-dimensional rectifier, while directly invoking Theorem~\ref{thm:convex-validity} would require analyzing a $d$-dimensional rectifier and thus yields more conservative intervals.

\begin{algorithm}[H]
\caption{\texttt{MeanCI} (see Proposition~\ref{prop:wsr-interval-iid})}
\label{alg:meanCI}
\begin{algorithmic}[1]
\Require data points $\{Z_1,\dots,Z_n\}$, error level $\alpha \in(0,1)$, range $[L,U]$ s.t. $Z_i\in[L,U]$ 
\State For all $i\in[n]$, let $Z_i \leftarrow (Z_i - L)/(U-L)$ \Comment{normalize data to interval $[0,1]$}
\State Construct fine grid $M_{\rm grid}$ of interval $[0,1]$
\State Initialize active set $\mathcal A = M_{\rm grid}$
\For{$t \in 1,\dots,n$}
\State Set $\hat{\mu}_t \leftarrow \frac{0.5 + \sum_{j=1}^t Z_{j}}{t+1}$, $\hat{\sigma}_t^2 \leftarrow \frac{0.25 + \sum_{j=1}^t (Z_{j} - \hat{\mu}_t)^2}{t + 1}$, $\lambda_t \leftarrow \sqrt{\frac{2\log(2/\alpha)}{n\hat{\sigma}^2_{t-1}}}$
\For{$m \in \mathcal A$}
\State $M_t^+(m)\leftarrow \left(1+\min\left(\lambda_t, \frac{0.5}{m}\right)(Z_{t} - m)\right) M^+_{t-1}(m)$
\State $M_t^-(m) \leftarrow \left(1-\min\left(\lambda_t, \frac{0.5}{1-m}\right)(Z_{t} - m)\right) M^-_{t-1}(m)$
\State $M_t(m) \leftarrow \frac{1}{2} \max\left\{M_t^+(m), \; M_t^-(m)   \right\}$ \Comment{construct test martingale for $m\in[0,1]$}
\If{$M_t(m) \geq 1/\alpha$}\\
\qquad \qquad \quad  $\mathcal A \leftarrow \mathcal A \setminus \{m\}$ \Comment{Remove $m$ from active set}
\EndIf
\EndFor
\EndFor
\Ensure 
Confidence set for the mean $\C_\alpha = \{m(U-L) + L: m\in \mathcal A\}$
\end{algorithmic}
\end{algorithm}

\subsection{Regularity conditions}
\label{app:regularity-conditions}

All algorithms stated in Section \ref{sec:convex} rely on confidence intervals derived from the central limit theorem. For such intervals to be asymptotically valid, we require that the two quantities whose mean is being estimated, namely $g_\theta(X_i,Y_i) - g_\theta(X_i,f(X_i))$ and $g_\theta(X_i,f(X_i))$, have at least the first two moments (see Proposition \ref{prop:clt-interval-asymptotic}).

For Proposition \ref{prop:ols} to hold, we need the same conditions as those required for classical linear regression intervals to cover the target. We note that these conditions are very weak; in particular, it is \emph{not} required that the true data-generating process be linear or the errors be homoskedastic. See Buja et al.~\cite{buja2019models} for a detailed discussion. The following are the required conditions, as stated in Theorem 3 of Halbert White's seminal paper \cite{white1980using}. The data $(X_1,Y_1),\dots,(X_n,Y_n)$ is generated as $X_i = h(Z_i)$, $Y_i = g(Z_i) + \epsilon_i$, where $(Z_i,\epsilon_i)$ are mean-zero i.i.d. random draws from some distribution such that $\E[Z_i Z_i^\top]$ and $\E[X_i X_i^\top]$ are finite and nonsingular, and $\E[\epsilon_i^2]$, $\E[Y_i^2 X_i X_i^\top]$, and $\E[X_{ij}^2 X_i X_i^\top]$ are all finite. In addition, we assume that $h$ and $g$ are measurable. Under these conditions,
$$\sqrt{n} (\hat \theta_{\rm OLS} - \theta^*) \Rightarrow \mathcal{N}(0,\Sigma^{-1}V\Sigma^{-1}),$$
where $\theta^* = \argmin_\theta \E [(Y_i - X_i^\top \theta)^2]$, $\hat \theta_{\rm OLS} = \argmin_\theta \frac{1}{n} \sum_{i=1}^n (Y_i - X_i^\top \theta)^2$, $\Sigma = \E[X_i X_i^\top]$, $V = \E[(Y_i - X_i^\top \theta^*)^2 X_i X_i^\top]$. Moreover, $\frac{1}{n} X^\top X \rightarrow \Sigma$ and $\frac{1}{n} \sum_{i=1}^n (Y_i - X_i^\top \hat\theta_{\rm OLS})^2 X_i X_i^\top \rightarrow V$ almost surely.

\section{Proofs}
\label{app:proofs}

\subsection{Proof of Theorem~\ref{thm:convex-validity}}

We show that $\theta^*\in\CPP_\alpha$ with probability at least $1-\alpha$; that is, with probability at least $1-\alpha$ it holds that
$$0\in\cR_\delta(\theta^*) + \T_{\alpha-\delta}(\theta^*).$$
Consider the event $E = \{\rec_{\theta^*} \in \cR_\delta(\theta^*) \} \cap \left\{\E [g_{\theta^*}(X_i, f(X_i))] \in \T_{\alpha-\delta}(\theta^*)\right\}$.
By a union bound, $P(E)\geq 1-\alpha$.
On the event $E$, we have that
\begin{align}
\E [g_{\theta^*}(X_i, Y_i)] &=  \E [g_{\theta^*}(X_i, Y_i)] - \E [g_{\theta^*}(X_i, f(X_i))] + \E[g_{\theta^*}(X_i, f(X_i))]\\
&= \rec_{\theta^*} + \E[g_{\theta^*}(X_i, f(X_i))] \in \cR_\delta(\theta^*) + \T_{\alpha-\delta}(\theta^*).
\end{align}
The theorem finally follows by invoking the nondegeneracy condition, which ensures $\E [g_{\theta^*}(X_i, Y_i)] = 0$, so we have shown $0\in\cR_\delta(\theta^*) + \T_{\alpha-\delta}(\theta^*)$.

\subsection{Proof of Theorem \ref{thm:convex-validity_asymptotic}}

We show that $\theta^*\not\in\CPP_\alpha$ with probability at most $\alpha$ in the limit; that is,
$$\limsup_{n,N\rightarrow\infty} \; P\left(\left|\rechat_{\theta^*,j} + \hat g^f_{\theta^*, j} \right|> z_{1-\alpha/(2p)}\sqrt{\frac{\hat\sigma^2_{\Delta,j}(\theta^*)}{n} + \frac{\hat\sigma^2_{g,j}(\theta^*)}{N}},~\forall j\in[p] \right) \leq \alpha.$$
For each $j\in[p]$, the central limit theorem implies that
$$\sqrt{n}(\rechat_{\theta^*,j} - \E[\rechat_{\theta^*,j}]) \Rightarrow \mathcal N(0,\sigma_{\Delta,j}^2(\theta^*));\quad \sqrt{N}(\hat g^f_{\theta^*, j} - \E[\hat g^f_{\theta^*, j}]) \Rightarrow \mathcal N(0,\sigma_{g,j}^2(\theta^*)),$$
where $\sigma_{\Delta,j}^2(\theta^*)$ is the variance of $g_{\theta^*,j}(X_i,Y_i) - g_{\theta^*,j}(X_i,f(X_i))$ and $\sigma_{g,j}^2(\theta^*)$ is the variance of $g_{\theta^*,j}(X_i,f(X_i))$. Therefore, by Slutsky's theorem, we get
\begin{align*}
\sqrt{N}(\rechat_{\theta^*,j} + \hat g^f_{\theta^*, j} - \E[\rechat_{\theta^*,j} + \hat g^f_{\theta^*, j}]) &= \sqrt{n}(\rechat_{\theta^*,j} - \E[\rechat_{\theta^*,j}])\sqrt{\frac{N}{n}} + \sqrt{N}( \hat g^f_{\theta^*, j} - \E[\hat g^f_{\theta^*, j}])\\
&\Rightarrow \mathcal N\left(0,\frac{1}{p}\sigma_{\Delta,j}^2(\theta^*)+\sigma_{g,j}^2(\theta^*)\right).
\end{align*}
This in turn implies
\begin{equation}
\label{eq:clt-implication}
\limsup_{n,N\rightarrow\infty} \; P\left(\left|\rechat_{\theta^*,j} + \hat g_j^f(\theta^*) - \E\left[\rechat_{\theta^*,j} + \hat g_j^f(\theta^*)\right] \right|> z_{1-\alpha/(2p)} \frac{\hat\sigma_j}{\sqrt{N}} \right) \leq \frac{\alpha}{p},
\end{equation}
where $\hat\sigma_j^2$ is a consistent estimate of the variance $\frac{1}{p}\sigma_{\Delta,j}^2(\theta^*)+\sigma_{g,j}^2(\theta^*)$. We take $\hat\sigma_j^2 = \hat\sigma_{\Delta,j}^2(\theta^*)\frac{N}{n} + \hat\sigma_{g,j}^2(\theta^*)$; this estimate is consistent since the two terms are individually consistent estimates of the respective variances.
Now notice that
\begin{equation}
\label{eq:grad0}
\E\left[\rechat_{\theta^*} + \hat g^f_{\theta^*}\right] = \E\left[g_{\theta^*}(X_i,Y_i)-g_{\theta^*}(X_i,f(X_i)) + g_{\theta^*}(\Xt_i,f(\Xt_i))\right] = \E[g_{\theta^*}(X_i,Y_i)] = 0,
\end{equation}
where the last step follows by the nondegeneracy condition. Putting together \eqref{eq:clt-implication}, \eqref{eq:grad0}, and the choice of $\hat\sigma_j$ derived above, and applying a union bound, we get
\begin{align*}
&\limsup_{n,N\rightarrow\infty} \; P\left(\exists  j\in[p]: \left|\rechat_{\theta^*,j} + \hat g_j^f(\theta^*) \right|> z_{1-\alpha/(2p)}\sqrt{\frac{\hat\sigma^2_{\Delta,j}(\theta^*)}{n} + \frac{\hat\sigma^2_{g,j}(\theta^*)}{N}} \right)\\
&\quad \leq \sum_{j=1}^p \limsup_{n,N\rightarrow\infty} P\left(\left|\rechat_{\theta^*,j} + \hat g_j^f(\theta^*) \right|> z_{1-\alpha/(2p)}\sqrt{\frac{\hat\sigma^2_{\Delta,j}(\theta^*)}{n} + \frac{\hat\sigma^2_{g,j}(\theta^*)}{N}} \right)\\
&\quad = \sum_{j=1}^p \limsup_{n,N\rightarrow\infty} P\left(\left|\rechat_{\theta^*,j} + \hat g_j^f(\theta^*) -  \E\left[\rechat_{\theta^*,j} + \hat g_j^f(\theta^*)\right]\right|> z_{1-\alpha/(2p)}\hat\sigma_j \right)\\
&\quad \leq \sum_{j=1}^p\frac{\alpha}{p}\\
&\quad = \alpha.
\end{align*}

\subsection{Proof of Proposition \ref{prop:mean_est_validity}}

We show that the prediction-powered confidence set constructed in Algorithm \ref{alg:means} is a special case of the prediction-powered confidence set constructed in Theorem \ref{thm:convex-validity_asymptotic}. The proof then follows directly by the guarantee of Theorem \ref{thm:convex-validity_asymptotic}.

Since $g_\theta(y) = \theta - y$, we have
$$\rechat_{\theta} \equiv \rechat = \frac 1 n \sum_{i=1}^n (f(X_i) - Y_i);\quad \hat g^f_{\theta} = \theta - \frac 1 N \sum_{i=1}^N f(\Xt_i).$$
Therefore, the set $\CPP_\alpha$ from Theorem \ref{thm:convex-validity_asymptotic} can be written as
$$\CPP_\alpha = \left\{\theta: \left|\theta - \frac 1 N \sum_{i=1}^N f(\Xt_i) + \frac 1 n \sum_{i=1}^n (f(X_i) - Y_i)\right| \leq w_\alpha(\theta)\right\} = \left(\frac 1 N \sum_{i=1}^N f(\Xt_i) - \frac 1 n \sum_{i=1}^n (f(X_i) - Y_i) \pm w_\alpha(\theta)\right).$$
This is exactly the set constructed in Algorithm \ref{alg:means}, which completes the proof.

\subsection{Proof of Proposition \ref{prop:quantile}}

Like in the proof of Proposition \ref{prop:mean_est_validity}, we proceed by showing that the prediction-powered confidence set constructed in Algorithm \ref{alg:quantile} is a special case of the prediction-powered confidence set constructed in Theorem~\ref{thm:convex-validity_asymptotic}. Then, we simply invoke Theorem \ref{thm:convex-validity_asymptotic}.

Since $g_\theta(y) = -q + \ind{y \leq \theta}$, we have
$$\rechat_{\theta} = \frac{1}{n} \sum_{i=1}^n \left(\ind{Y_i \leq \theta} - \ind{f(X_i) \leq \theta} \right);\quad \hat g^f_{\theta} = -q + \hat F(\theta),$$
where $\hat F(\theta) = \frac{1}{N} \sum_{i=1}^N \ind{f(\Xt_i) \leq \theta}$.
Therefore, the set $\CPP_\alpha$ from Theorem \ref{thm:convex-validity_asymptotic} can be written as
$$\CPP_\alpha = \left\{\theta: \left|\frac{1}{n} \sum_{i=1}^n \left(\ind{Y_i \leq \theta} - \ind{f(X_i) \leq \theta} \right) - q + \hat F(\theta)\right| \leq w_\alpha(\theta)\right\} = \left\{\theta: \left|\hat F(\theta)+\rechat_{\theta} - q\right| \leq w_\alpha(\theta)\right\}.$$
This is exactly the set constructed in Algorithm \ref{alg:quantile}. Therefore, the guarantee of Proposition \ref{prop:quantile} follows by the guarantee of Theorem \ref{thm:convex-validity_asymptotic}.

\subsection{Proof of Proposition \ref{prop:logistic}}

The proof follows a similar pattern as the previous two propositions, by arguing that the prediction-powered confidence set constructed in Algorithm \ref{alg:logistic} is a special case of the prediction-powered confidence set constructed in Theorem \ref{thm:convex-validity_asymptotic}.

Since $g_\theta(x,y) = x(\mu_\theta(x) - y)$, we have
$$\rechat_{\theta} \equiv \rechat = \frac 1 n \sum_{i=1}^n X_i (f(X_i) - Y_i);\quad \hat g^f_{\theta} = \frac{1}{N} \sum_{i=1}^N \Xt_i (\mu_\theta(\Xt_i) - f(\Xt_i)).$$
These quantities are explicitly computed in Algorithm \ref{alg:logistic}. Moreover, the set $\CPP_\alpha$ constructed in Algorithm \ref{alg:logistic} exactly follows the recipe of Theorem \ref{thm:convex-validity_asymptotic}, so the proof immediately follows.

\subsection{Proof of Proposition \ref{prop:ols}}

For linear regression, we can derive more powerful prediction-powered confidence intervals than those implied by Theorem \ref{thm:convex-validity} by exploiting the linearity of the least-squares estimator.

Recall that Theorem \ref{thm:convex-validity_asymptotic} assumes that $\frac{n}{N}\rightarrow p$, for some fraction $p\in(0,1)$.

Theorem 3 of White~\cite{white1980heteroskedasticity} implies that
$$\sqrt{n}(\rechat - \rec) \Rightarrow \mathcal{N}(0, W);\quad \sqrt{N}(\thetaf - \thetaflimit) \Rightarrow \mathcal{N}(0, W'),$$
for appropriately defined coviariance matrices $W$ and $W'$, where $\thetaflimit = (\E[X_i X_i^\top])^{-1} \E[X_i f(X_i)]$ and $\rec = (\E[X_i X_i^\top])^{-1} \E[X_i (f(X_i)-Y_i)]$. With this, we can write the target estimand as $\theta^* = (\E[X_i X_i^\top])^{-1} \E[X_i Y_i] = \thetaflimit - \rec$.

Combining Theorem 3 of White with Slutsky's theorem, we get
$$\sqrt{N}(\thetaPP - \theta^*) = \sqrt{N}(\thetaf - \thetaflimit) - \sqrt{n}(\rechat - \rec)\sqrt{\frac{N}{n}} \Rightarrow \mathcal N\left(0, W\frac{1}{p} + W' \right).$$
White also shows that $V$ and $\tilde V$, as defined in Algorithm \ref{alg:ols}, are consistent estimates of $W$ and $W'$, respectively. Therefore, $\thetaPP$ is asymptotically normal and consistent, and we have a consistent estimate of its covariance. In particular,
$$V_{j^*j^*}\frac{N}{n} + \tilde V_{j^*j^*} \rightarrow W_{j^*j^*}\frac{1}{p} + W'_{j^*j^*}.$$
This means that we can construct asymptotically valid confidence intervals via a normal approximation by choosing width $z_{1-\alpha/2}\sqrt{V_{j^*j^*}\frac{N}{n} + \tilde V_{j^*j^*}} \sqrt{\frac{1}{N}} = z_{1-\alpha/2}\sqrt{\frac{V_{j^*j^*}}{n} + \frac{\tilde V_{j^*j^*}}{N}}$, and this is precisely what Algorithm~\ref{alg:ols} accomplishes.

\subsection{Proof of Theorem \ref{thm:m-estimators}}

Define
$$L(\theta) = \E [\ell_\theta(X_i,Y_i)], \quad L^f(\theta) = \E [\ell_\theta(X_i,f(X_i))].$$
By the definition of $\theta^*$, we have
\begin{align*}
\tilde L^f(\theta^*) &= (\tilde L^f(\theta^*) -  L(\theta^*)) + ( L(\theta^* ) - L(\thetaf) ) +  (  L(\thetaf) - \tilde L^f(\thetaf))  + \tilde L^f(\thetaf) \\
&\leq (\tilde L^f(\theta^*) -  L(\theta^* )) +  (  L(\thetaf) - \tilde L^f(\thetaf))  + \tilde L^f(\thetaf).
\end{align*}
By applying the validity of the confidence bounds, a union bound implies that with probability $1-\alpha$ we have
\begin{align*}
\tilde L^f(\theta^*) &\leq ( L^f(\theta^*) -  L(\theta^* )) +  (  L(\thetaf) -  L^f(\thetaf))  + \tilde L^f(\thetaf) + \T_{\frac{\alpha-\delta}{2}}^u(\theta^*) - \T_{\frac{\alpha-\delta}{2}}^l(\thetaf) \\
&= - \rec_{\theta^*} + \rec_{\thetaf} + \tilde L^f(\thetaf) + \T_{\frac{\alpha-\delta}{2}}^u(\theta^*) - \T_{\frac{\alpha-\delta}{2}}^l(\thetaf)   \\
&\leq -\cR_{\delta/2}^l(\theta^*) + \cR_{\delta/2}^u(\thetaf) + \tilde L^f(\thetaf) + \T_{\frac{\alpha-\delta}{2}}^u(\theta^*) - \T_{\frac{\alpha-\delta}{2}}^l(\thetaf). \\
\end{align*}
Therefore, with probability $1-\alpha$ we have that $\theta^* \in \CPP_\alpha$, as desired.

\subsection{Proof of Theorem~\ref{thm:label-shift}}
\label{app:label-shift}

Notice that we can write $\E_{\Q_Y} [\nu(Y)] = \nu^\top \Q_Y$, where on the right-hand side we are treating $\nu = (\nu(1),\dots,\nu(K))$ and $\Q_Y = (\Q_Y(1),\dots,\Q_Y(K))$ as vectors of length $K$. We can write similar expressions for $\Q_f,\widehat\Q_Y$, etc. Using this notation, by triangle inequality we have
\begin{equation}
\label{eq:triangle-label-shift}
|\theta^*- \nu^\top \widehat\Q_Y| =
|\nu^\top \Q_Y- \nu^\top \widehat\Q_Y| \leq \left|\nu^\top \widehat{\K}^{-1}(\Q_f - \widehat \Q_f) \right| + \left| \nu^\top\K^{-1}\Q_f - \nu^\top \widehat{\K}^{-1}\Q_f \right|.
\end{equation}
We bound the first term using H\"older's inequality,
\begin{equation}
\left|\nu^\top \widehat{\K}^{-1}(\Q_f - \widehat \Q_f) \right| \leq \|\nu^\top \widehat{\K}^{-1}\|_1 \|\Q_f - \widehat\Q_f\|_\infty. 
\end{equation}
For the second term, we write
\begin{equation}
\left| \nu^\top\K^{-1}\Q_f - \nu^\top \widehat{\K}^{-1}\Q_f \right| = \left| \nu^\top \widehat{\K}^{-1}(\widehat{\K}-\K) \K^{-1}\Q_f\right|.
\end{equation}
In the above equation, the factor on the right, $\K^{-1}\Q_f$, is exactly equal to $\Q_Y$, and thus lives on the simplex, which we denote by $\Delta$.
Using this fact and H\"older's inequality,
\begin{equation}
\label{eq:holders-general-label-shift}
\left| \nu^\top \widehat{\K}^{-1}(\widehat{\K}-\K) \K^{-1}\Q_f\right| \leq \sup_{q \in \Delta} \left| \nu^\top \widehat{\K}^{-1}(\widehat{\K}-\K) q \right| \leq \left\| \nu^\top \widehat{\K}^{-1}\right\|_1 \sup_{q \in \Delta} \left\| (\widehat{\K}-\K)q \right\|_\infty.
\end{equation}
Next, we have
\begin{equation}
\sup_{q \in \Delta} \|(\widehat \K-\K)q\|_\infty = \max_{k\in[K]} \|\widehat \K_k - \K_k\|_\infty,
\end{equation}
where $\K_k$ indexes the $k$-th column of $\K$. This yields the expression
\begin{equation}
        \left\| \nu^\top \widehat{\K}^{-1}\right\|_1 \sup_{q\in\Delta}\left\| (\widehat{\K}-\K)q \right\|_\infty =  \left\| \nu^\top \widehat{\K}^{-1}\right\|_1 \max_{k\in[K]} \|\widehat \K_k - \K_k\|_\infty.
\end{equation}
Putting everything together and going back to \eqref{eq:triangle-label-shift}, we have
\begin{equation}
\label{eq:deterministic-endpoint-labelshift}
|\nu^\top \Q_Y- \nu^\top \widehat\Q_Y| \leq \|\nu^\top \widehat{\K}^{-1}\|_1\left(\|\Q_f - \widehat\Q_f\|_\infty + \max_{k\in[K]} \|\widehat{\K}_k-\K_k \|_\infty\right).
\end{equation}
Since $\|\nu^\top \widehat{\K}^{-1}\|_1$ can be evaluated empirically, it remains to bound the distributional distances $\|\Q_f - \widehat\Q_f\|_\infty$ and $\max_{k\in[K]} \|\widehat{\K}_k-\K_k \|_\infty$.

For the first term, we can simply apply the DKWM inequality \cite{dvoretzky1956asymptotic,massart1990tight}, which gives
\begin{equation}
\label{eq:bound-qhat-labelshift}
\|\Q_f - \widehat\Q_f\|_\infty \leq \sqrt{ \frac{2}{N}\log \frac{2}{\alpha-\delta}}
\end{equation}
with probability $1-(\alpha-\delta)$.
See \cite{canonne2020short} for details.

For the second term, $\max_{k\in[K]} \|\widehat{\K}_k-\K_k \|_\infty$, since we only have $n$ observations for estimation, we use a more adaptive concentration result. In particular, for each $l,k\in[K]$, $n(k)\widehat{\K}_{l,k}$ (conditional on the $k$-th column) follows a binomial distribution with $n(k)$ draws and success probability $\K_{l,k}$. Therefore, if we let
$$C_{l,k} = \left\{p: n(k) \widehat \K_{l,k} \in \left( F^{-1}_{\mathrm{Binom}(n(k),p)}\left(\frac{\delta}{2K^2}\right), F^{-1}_{\mathrm{Binom}(n(k),p)}\left(1-\frac{\delta}{2K^2}\right) \right) \right\},$$
where $F_{\mathrm{Binom}(n(k),p)}$ denotes the Binomial CDF, then by a union bound:
\begin{equation}
\label{eq:bound-kmax-labelshift}
P\left(\max_{k\in[K]} \|\widehat \K_k - \K_k\|_\infty \geq \max_{l,k\in[K]} \max_{p\in C_{l,k}} |\widehat \K_{l,k} - p|\right) \leq \delta.
\end{equation}
Combining equations~\eqref{eq:deterministic-endpoint-labelshift},~\eqref{eq:bound-qhat-labelshift} and~\eqref{eq:bound-kmax-labelshift} yields the final result.

\subsection{Proof of Corollary \ref{cor:means_cor}}

The proof follows by instantiating the terms in Theorem \ref{thm:convex-validity}. In particular, we have $\E[g_\theta(f(X_i))] = \theta - \E[f(X_i)]$, hence it is valid to construct $\T_{\alpha-\delta}(\theta)$ as:
$$\E[g_\theta(f(X_i))] \in \T_{\alpha-\delta}(\theta) = \theta -  (f_{\alpha-\delta}^l, f_{\alpha-\delta}^u ).$$
Therefore, the condition $0\in \cR_\delta + \T_{\alpha-\delta}(\theta)$ becomes 
$$0 \in (\cR_\delta^l,\cR_\delta^u) + \theta - (f_{\alpha-\delta}^l, f_{\alpha-\delta}^u),$$
which after rearranging and simplifying is equivalent to
$$\theta\in \left(f_{\alpha-\delta}^l - \cR_\delta^u, f_{\alpha-\delta}^u - \cR_\delta^l\right).$$
This set exactly matches the set $\CPP_\alpha$ constructed in Algorithm \ref{alg:means_nonasymp}.

\subsection{Proof of Corollary \ref{cor:quantile_cor}}

The proof follows by instantiating the terms in Theorem \ref{thm:convex-validity}. First, we have $\E[g_\theta(f(X_i))] = -q + P(f(X_i)\leq \theta)$; therefore, it is valid to construct $\T_{\alpha-\delta}(\theta)$ as:
$$\E[g_\theta(f(X_i))] \in \T_{\alpha-\delta}(\theta) = -q + \left(\hat F^l_{\alpha-\delta}(\theta),  \hat F^u_{\alpha-\delta}(\theta)\right).$$
Therefore, the condition $0\in \cR_\delta(\theta) + \T_{\alpha-\delta}(\theta)$ becomes 
$$q \in \left(\hat F^l_{\alpha-\delta}(\theta) + \cR_\delta^l(\theta), \hat F^u_{\alpha-\delta}(\theta) + \cR_\delta^u(\theta) \right),$$
which matches the condition used to form $\CPP_\alpha$ in Algorithm \ref{alg:quantile_nonasymp}.

\subsection{Proof of Corollary \ref{cor:logistic_valid}}

We instantiate the relevant terms in Theorem \ref{thm:convex-validity}. We have $\E[g_\theta(X_i,f(X_i))] = \E\left[-X_i f(X_i) + X_i \frac{1}{1 + \exp(-X_i^\top\theta)}\right]$. Note that, because $X_i$ is coordinatewise bounded, and $Y_i,\frac{1}{1 + \exp(-X_i^\top\theta)}\in[0,1]$, we have $|(g_\theta(X_i,f(X_i)))_j| \leq B_j$ almost surely. Therefore, we can construct $\T_{\alpha-\delta}(\theta)$ as:
$$\E[g_\theta(X_i,f(X_i))] \in \T_{\alpha-\delta}(\theta) = \left( g_{\alpha-\delta}^l(\theta), g_{\alpha-\delta}^u(\theta) \right) = \left( g_{\alpha-\delta,1}^l(\theta), g_{\alpha-\delta,1}^u(\theta) \right) \times \dots \times \left( g_{\alpha-\delta,d}^l(\theta), g_{\alpha-\delta, d}^u(\theta) \right).$$
Since the rectifier has no dependence on $\theta$, the condition $0\in \cR_\delta(\theta) + \T_{\alpha-\delta}(\theta)$ becomes 
$$0 \in (\cR_{\delta,j}^l,\cR_{\delta,j}^u) + \left( g_{\alpha-\delta,j}^l(\theta), g_{\alpha-\delta,j}^u(\theta) \right),\quad \forall j\in[d],$$
which matches the condition in $\CPP_\alpha$ in Algorithm \ref{alg:logistic_nonasymp}.

\section{Confidence intervals for the mean}
\label{app:cis}

We give an overview of off-the-shelf confidence intervals for the mean. We state the results for two observation models: first for the i.i.d. sampling model considered in the main body and then for the finite-population setting discussed in Appendix \ref{app:finite-pop}. In both cases, we provide a construction with nonasymptotic guarantees and one with asymptotic guarantees.

For the nonasymptotic confidence intervals, we rely on the results of Waudby-Smith and Ramdas~\cite{waudby2020variance}, specifically their Theorem 3 and Theorem 4. We opt for these results because of their strong practical performance, which is primarily driven by variance adaptivity. These results assume that the observed random variables are bounded within a known interval. Without loss of generality we assume that the observations are bounded in $[0,1]$ (otherwise we can always normalize the observations to $[0,1]$).

For the asymptotic confidence intervals, we rely on the central limit theorem (CLT) and its variant for sampling without replacement; see \cite{erdos1959central,hoglund1978sampling} for classical references.

\subsection{Inference with i.i.d. samples}

In the following two results, assume that we observe $Z_1, \dots, Z_n
\stackrel{\rm i.i.d.}{\sim} \P$ and let $\mu = \E[Z_i]$.

\begin{prop}[Nonasymptotic CI: Theorem 3 in \cite{waudby2020variance}]
\label{prop:wsr-interval-iid}
Assume $\text{supp}(\P)\subseteq [0,1]$. Let
\begin{equation}
    \hat{\mu}_t = \frac{0.5 + \sum_{j=1}^t Z_{j}}{t+1},\quad \hat{\sigma}_t^2 = \frac{0.25 + \sum_{j=1}^t (Z_{j} - \hat{\mu}_t)^2}{t + 1}, \quad \lambda_t = \sqrt{\frac{2\log(2/\alpha)}{n\hat{\sigma}^2_{t-1}}}.
\end{equation}
For every $m\in[0,1]$, define the supermartingale:
\begin{equation}
    M_t(m) = \frac{1}{2} \max\left\{\prod\limits_{j=1}^t\left(1+\min\left(\lambda_j, \frac{0.5}{m}\right)(Z_{j} - m)\right), \;  \prod\limits_{j=1}^t\left(1-\min\left(\lambda_j, \frac{0.5}{1-m}\right)(Z_{j} - m)\right) \right\}.
\end{equation}
Let
\begin{equation}
    \C = \bigcap\limits_{t=1}^{n} \left\{ m \in [0,1] : M_t(m) < 1/\alpha \right\}.
\end{equation}
Then,
$$P\left( \mu \in \C \right) \geq 1-\alpha.$$
\end{prop}

Intuitively, the supermartingale $M_t(m)$ should be thought of as the amount of evidence against $m$ being the true mean. That is, $M_t(m)$ being big suggests that $m$ is unlikely to be the true mean, so the final confidence set is the collection of all $m$ for which the amount of such evidence is small.

For large $n$, computing the intersection in the definition of $\C$ can be intractable, so we conservatively choose a subsequence of $1, \dots, n$ for the computation.

\begin{prop}[Asymptotic CI: CLT interval]
\label{prop:clt-interval-asymptotic}
Assume $\P$ has a finite second moment. Let
$$\C = \left(\frac{1}{n} \sum_{i=1}^n Z_i \pm z_{1-\alpha/2} \frac{\hat \sigma}{\sqrt{n}} \right),$$
where $\hat \sigma = \sqrt{\frac{1}{n}\sum_{i=1}^n (Z_i - \frac{1}{n}\sum_{j=1}^n Z_j)^2}$.
Then,
$$\liminf_{n\rightarrow\infty} P\left(\mu \in \C \right)\geq 1-\alpha.$$

\end{prop}

\subsection{Inference on a finite population}

In the following two results, we assume that there exists a \emph{fixed} sequence $Z_1,\dots,Z_N$, and we observe $\{Z_i:i\in\Ical\}$, where $\Ical = \{i_1,\dots,i_n\}$ is a uniform random subset of $[N]$ with cardinality $n$. We let $\mu = \frac{1}{N} \sum_{i=1}^N Z_i$. For the asymptotic result, we assume that $Z_1,\dots,Z_N$ is the first $N$ entries of an infinite underlying sequence $Z_1,Z_2,\dots$.

\begin{prop}[Nonasymptotic CI: Theorem 4 in \cite{waudby2020variance}]
\label{prop:wsr-finite-pop}
Assume $Z_i\in [0,1],~i\in[N]$. Let
\begin{equation}
    \hat{\mu}_t = \frac{0.5 + \sum_{j=1}^t Z_{i_j}}{t+1},\quad \hat{\sigma}_t^2 = \frac{0.25 + \sum_{j=1}^t (Z_{i_j} - \hat{\mu}_t)^2}{t + 1}, \quad \lambda_t = \sqrt{\frac{2\log(2/\alpha)}{n\hat{\sigma}^2_{t-1}}}.
\end{equation}
For every $m\in[0,1]$, define the supermartingale:
\begin{equation}
    M_t(m) = \frac{1}{2} \max\left\{\prod\limits_{j=1}^t\left(1+\min\left(\lambda_j, \frac{0.5}{\mu_t(m)}\right)(Z_{i_j} - \mu_t(m))\right), \;  \prod\limits_{j=1}^t\left(1-\min\left(\lambda_j, \frac{0.5}{1-\mu_t(m)}\right)(Z_{i_j} - \mu_t(m))\right) \right\},
\end{equation}
where $\mu_t(m) = \frac{Nm - \sum_{j=1}^{t-1}Z_{i_j}}{N - t + 1}$ is the putative mean. Let
\begin{equation}
    \C = \bigcap\limits_{t=1}^{n} \left\{ m \in [0,1] : M_t(m) < 1/\alpha \right\}.
\end{equation}
Then,
$$P\left( \mu \in \C \right) \geq 1-\alpha.$$
\end{prop}

\begin{prop}[Asymptotic CI: CLT for sampling without replacement]
\label{prop:finite-partial-sum-concentration}
Let $\sigma^2 = \frac 1 N \sum_{i=1}^N (Z_i - \mu)^2$, and $\hat{\sigma}^2 = \frac 1 n \sum_{i \in \Ical} (Z_i - \hat{\mu})^2$.
Assume that $\mu$ and $\sigma$ have a limit and that $n/N\rightarrow p$ for some $p\in(0,1)$. Let
$$\C = \left(\frac{1}{n}\sum\limits_{i \in \Ical} Z_i \pm z_{1-\alpha/2} \frac{\hat{\sigma}}{\sqrt{n}} \sqrt{\frac{N-n}{N}} \right).$$
Then,
\begin{equation}
    \liminf_{n,N\rightarrow\infty} P\left( \mu \in \C \right) \geq 1-\alpha.
\end{equation}
\end{prop}

\section{Comparison to baseline procedures}
\label{app:comparisons}

We compare prediction-powered inference with two baseline procedures that also combine labeled and unlabeled data in performing statistical inference. The baselines are:
\begin{enumerate}
    \item \textbf{Post-prediction inference.} We use the post-prediction inference procedure of Wang et al. \cite{wang2020methods} to estimate ordinary least-squares (OLS) coefficients. The procedure first fits a regression $r$ to predict $Y_i$ from $f(X_i)$ on the gold-standard dataset. Subsequently, the regression function is used to correct the imputed labels on the unlabeled dataset. Confidence intervals are formed using the $r(f(\Xt_i))$ as if they were gold-standard data. This procedure has no theoretical guarantees in general and requires strong distributional assumptions on the relationship between $Y_i$ and $f(X_i)$ to provide coverage. Our experiments indicate that this approach fails to cover in realistic conditions.
    \item \textbf{Semi-supervised mean estimation.} The semi-supervised mean estimation procedure of Zhang and Bradic \cite{zhang2022high} involves cross-fitting a (possibly-regularized) linear model on $K$ distinct folds of the gold-standard dataset. The average of the $K$ model predictions on each unlabeled data point is taken as its corresponding prediction $\hat Y$, and the average bias $\hat{Y}-Y$ of the $K$ models is also computed and used to debias the resulting mean estimate. The formal validity of this approach applies to mean estimation and requires the cross-fitting of linear models; it does not have formal guarantees for more flexible model classes. For this reason, it provides little improvement over the classical confidence interval in our experiments, since the variance reduction possible with linear models is typically limited.
\end{enumerate}

\subsection{Experimental protocol}

We evaluate the methods on an income prediction task on the same census dataset used for the logistic regression experiments in the main text. In the case of the semi-supervised baseline, the goal is to estimate the mean income in California in the year 2019 among employed individuals using a small amount of labeled data and a large amount of covariates. In the case of the post-prediction inference baseline, the target of inference is the OLS coefficient between age and income. The setup is the same as the logistic regression experiment described in the main text (including the use of the Folktables \cite{ding2021retiring} interface and the gradient-boosted tree \cite{chen2016xgboost} as the predictor).

\subsection{Comparison to post-prediction inference}

\begin{figure}[t]
    \centering
    \includegraphics[width=0.66\textwidth]{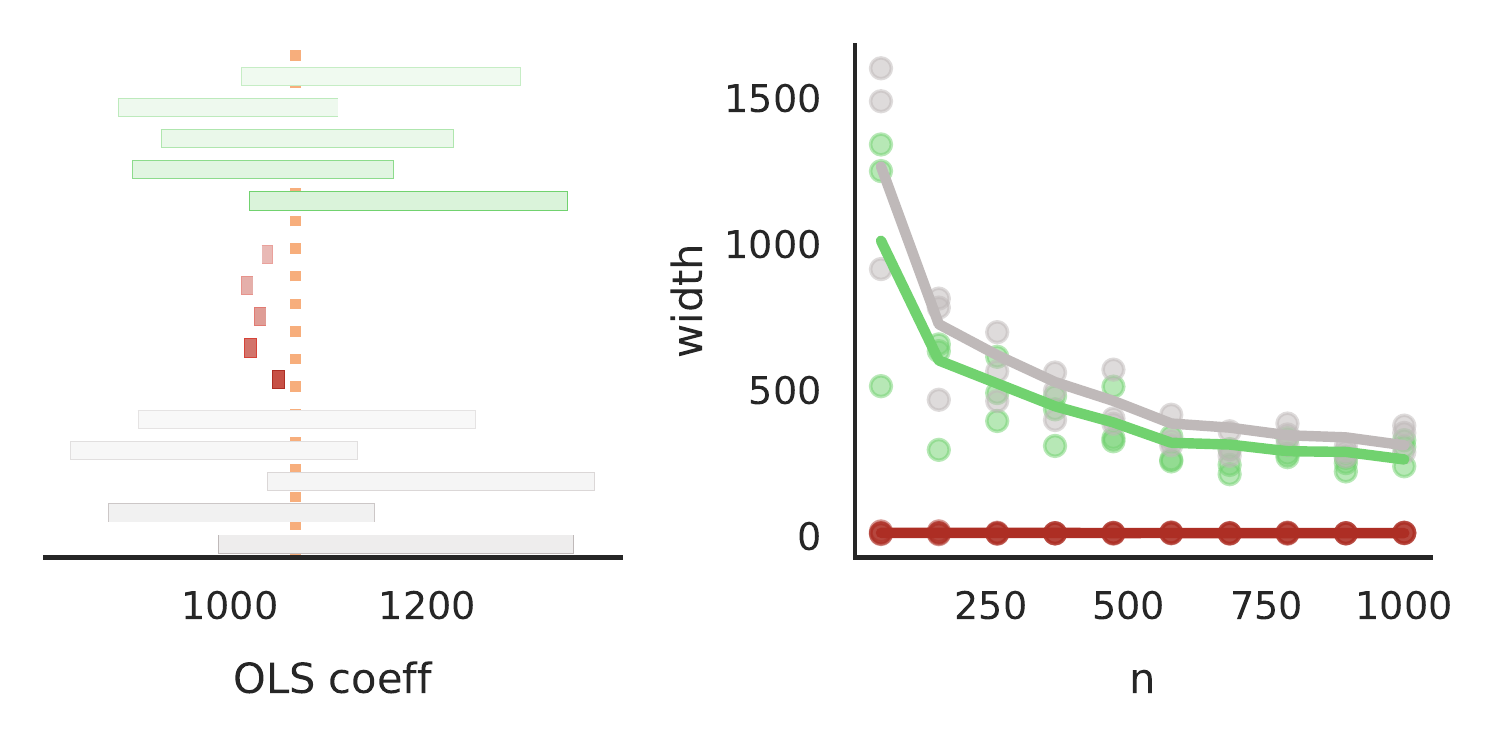}
    \caption{\textbf{Comparison to the post-prediction inference procedure.} On the left are five independent random draws of intervals with $n=1000$. On the right is a line plot of interval width as a function of $n$, averaged over $100$ independent trials. Five draws of interval widths are shown as a scatter plot at their respective $n$. The post-prediction inference approach is shown in red, the classical approach is in gray, and the prediction-powered approach is in green. The post-prediction inference approach has diminishing coverage in the experiment.}
    \label{fig:postprediction-comparison}
\end{figure}

Results of the post-prediction inference protocol as compared to the classical and prediction-powered approaches are shown in Figure  \ref{fig:postprediction-comparison} for the previously-described OLS coefficient between age and income.
The procedure does not cover at the proper rate and the intervals are biased.

\subsection{Comparison to semi-supervised mean estimation}

\begin{figure}[t]
    \centering
    \includegraphics[width=0.66\textwidth]{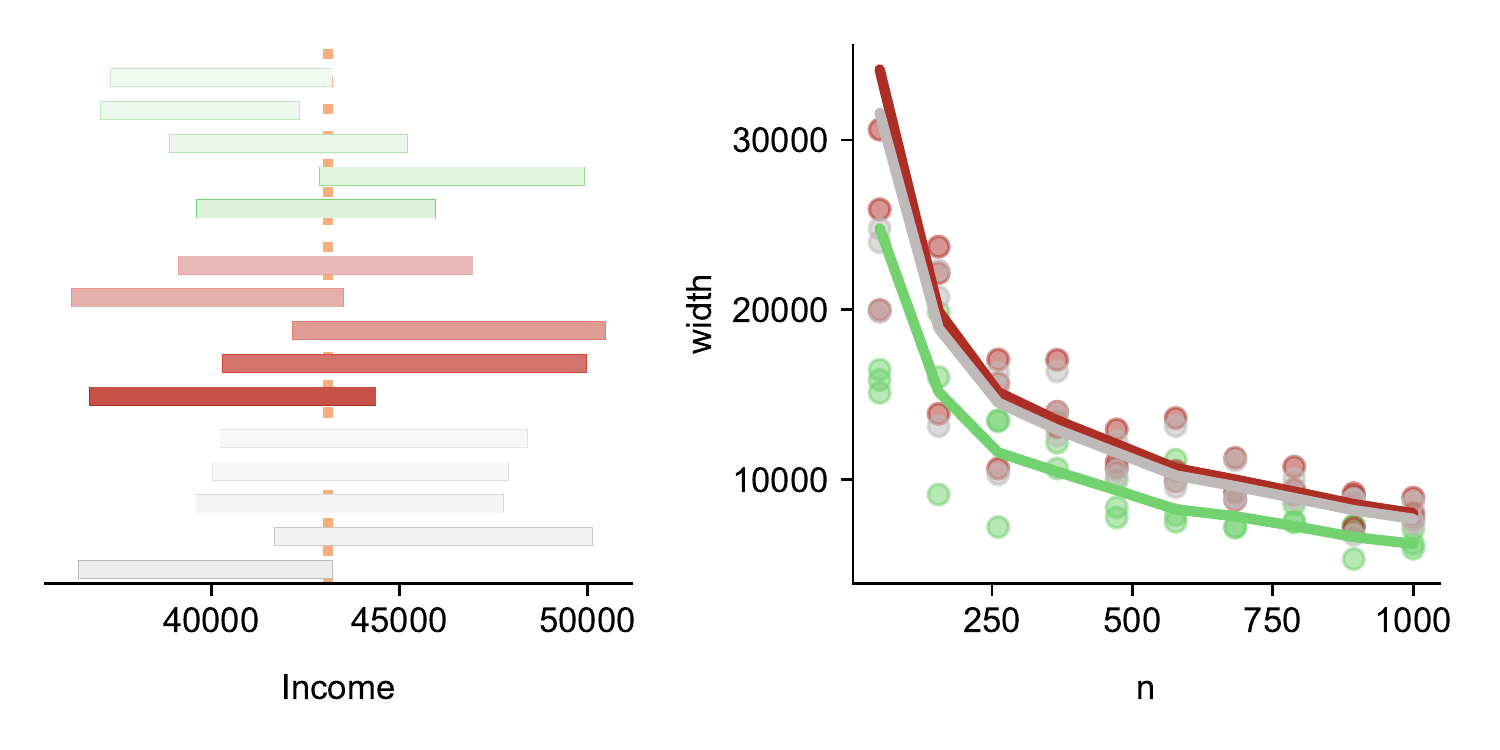}
    \caption{\textbf{Comparison to the semi-supervised mean estimation procedure.} The plot is the same as in Figure \ref{fig:postprediction-comparison}, but with semi-supervised inference shown in red. The semi-supervised intervals have a similar width to the classical ones in this experiment, while the prediction-powered intervals dominates.}
    \label{fig:semisupervised-comparison}
\end{figure}

Results of the semi-supervised mean estimation protocol as compared to the classical and prediction-powered approaches are shown in Figure \ref{fig:semisupervised-comparison} for the previously described mean income estimation task.
The prediction-powered intervals dominate both the semi-supervised intervals and the classical ones in the experiment for all values of $n$.

\section{Cases where prediction-powered inference is underpowered}
\label{sec:underpowered}

Since standard confidence intervals scale with the standard error of the estimator, prediction-powered inference is powerful when a machine-learning model can provide a reduction in the estimator variance. At a high level, this happens when $N$ is large enough relative to $n$ and the model is accurate enough. This was the case in all the experiments shown in the main text. This section precisely quantifies what it means to have an accurate enough model and large enough $N$. Corroborating the theory, we present two cases where classical inference outperforms prediction-powered inference: one where the model is not good  enough and another where $N$ is too small.

\subsection{Mathematical derivation}

Consider the case of mean estimation, $\theta^* = \E[Y_i]$. 
The widths of the classical confidence interval based on the central limit theorem and the prediction-powered confidence interval based on Algorithm \ref{alg:means} scale with $\Var(\thetaclass)$ and $\Var(\thetaPP)$, respectively, where $\thetaclass$ and $\thetaPP$ are defined in Section \ref{subsec:warmup}. The classical estimator has variance equal to
$$\Var(\thetaclass) = \frac{1}{n}\Var(Y_i).$$
The variance of the prediction-powered estimator equals
$$\Var(\thetaPP) = \frac{1}{N}\Var(f(X_i)) + \frac{1}{n}\Var(f(X_i) - Y_i).$$
Therefore, the prediction-powered confidence interval will be tighter when
$$\frac{1}{N}\Var(f(X_i)) + \frac{1}{n}\Var(f(X_i) - Y_i) < \frac 1 n \Var(Y_i).$$
Since the predictions $f(X_i)$ will typically have a variance that is of the same order as the variance of $Y_i$, if $N\approx n$ one should not expect prediction-powered inference to help. Gains are expected when $N\gg n$. In that case, $\frac{1}{N}\Var(f(X_i)) \ll \frac{1}{n}\Var(f(X_i) - Y_i)$, and thus prediction-powered inference helps when
$$\Var(f(X_i) - Y_i) <  \Var(Y_i).$$
In other words, prediction-powered inference gives tighter confidence intervals when the predictions explain away some of the outcome variance.

To gain further intuition, suppose that the outcomes are binary, $Y_i\sim \mathrm{Bern}(p)$, where $\mathrm{Bern}(p)$ denotes the Bernoulli distribution with parameter $p$. In this case, $\theta^* = p$. For simplicity, suppose that $P(f(X_i) = 0 | Y_i = 1) = P(f(X_i) = 1 | Y_i = 0) = \eta$. Then, a direct variance calculation gives $\Var(f(X_i) - Y_i) = \eta - \eta^2(1-2p)^2$ and $\Var(Y_i) = p(1-p)$. This allows for a direct comparison of the variances in terms of the outcome bias $p$ and model error $\eta$. For example, when $p=0.5$, the model error $\eta$ has to be smaller than $25\%$ for prediction-powered inference to yield smaller intervals; when $p=0.1$, meaning the outcomes themselves have low variance, the model error $\eta$ has to be smaller than about $9.5\%$. In general, the lower the variance of the outcome, the lower the model error has to be for prediction-powered inference to be helpful.

Putting everything together, the main takeaway is as follows: prediction-powered inference should only be applied when $N$ is (preferably substantially) larger than $n$, and when the model has a high enough predictive accuracy to explain away some of the outcome variance. While this derivation focused on mean estimation, a similar intuition holds for other estimation problems.

\subsection{Inaccurate machine-learning model}

\begin{figure}[ht]
    \centering
    \includegraphics[width=0.66\textwidth]{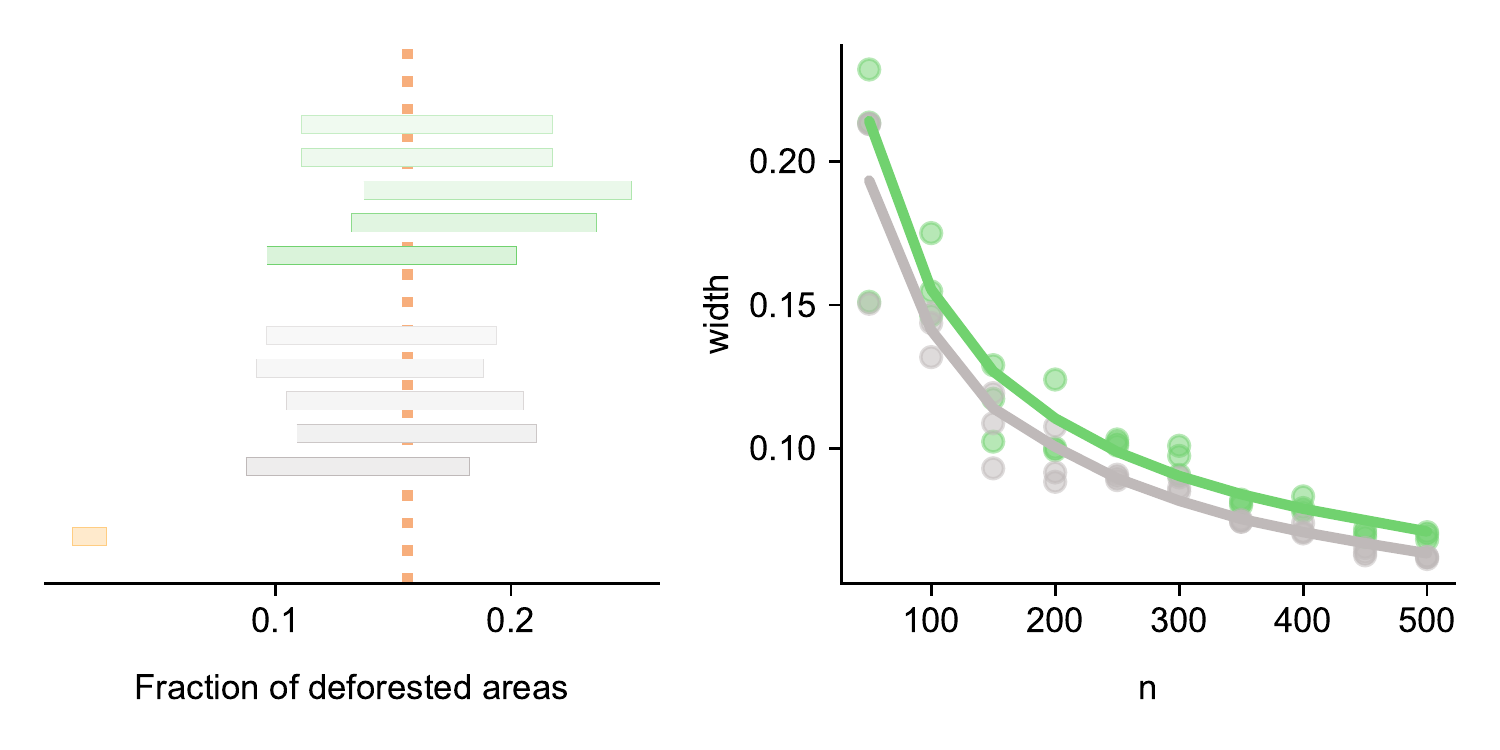}
    \caption{\textbf{Deforestation analysis with a linear model.} This is the same figure as Figure \ref{fig:big-figure}D, with the same color coding; the prediction-powered approach is green, the classical approach is gray, and the imputation approach is gold. However, the gradient-boosted tree is replaced with an ordinary linear regression. The drop in performance causes the classical intervals to outperform the prediction-powered intervals in terms of power.}
    \label{fig:bad-model-comparison}
\end{figure}

We repeat the deforestation analysis experiment from the main text. However, instead of a gradient-boosted tree, we use a linear regression model for prediction. This degrades predictive performance enough that the classical baseline outperforms the prediction-powered approach. See Figure \ref{fig:bad-model-comparison} for the results. Due to the reduction of power, for the same null hypothesis tested in the main text, the prediction-powered approach requires $n=40$ data points to reject, while the classical baseline requires $n=35$.

\subsection{Unlabeled dataset is too small}

\begin{figure}[ht]
    \centering
    \includegraphics[width=0.66\textwidth]{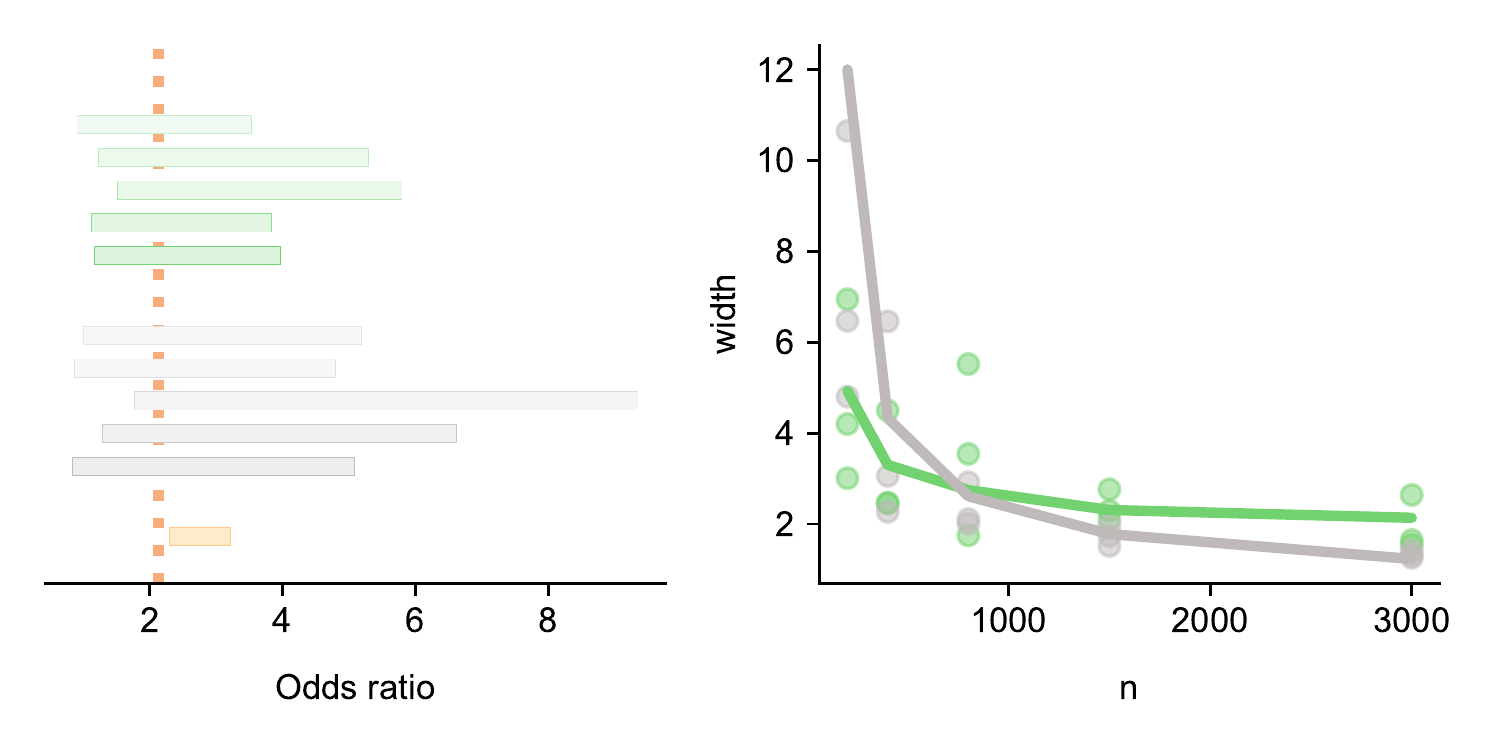}
    \caption{\textbf{AlphaFold analysis with a small unlabeled dataset.} This is the same figure as Figure \ref{fig:big-figure}A, with the same color coding; the prediction-powered approach is green, the classical approach is gray, and the imputation approach is gold. However, here $N$ is taken to be $1000$. It can be seen that, when $n > N$, the classical baseline outperforms the prediction-powered one.}
    \label{fig:lowN-comparison}
\end{figure}

We repeat the AlphaFold-based proteomic analysis from the main text. However, $N=1000$ data points are randomly chosen as the unlabeled dataset. The rest of the procedure is performed exactly the same way as described in the main text. 
The decrease in the unlabeled sample size leads to a reduction in power, and in the regime $n > N$, the classical baseline outperforms the prediction-powered approach. See Figure \ref{fig:lowN-comparison} for the results. For the same null hypothesis as in the main text, the prediction-powered approach requires $n=869$ data points to reject, while the classical baseline requires $n=652$.

\section{Experimental particulars}

\subsection{Relating protein structure and post-translational modifications}
\label{app:proteins}

The predictive model of whether a sequence position is in an intrinsically disordered region (IDR), $f$, is a logistic regression model that maps the relative solvent-accessible surface area (RSA) of each position, computed based on the AlphaFold-predicted structure using Bio.PDB \cite{hamelryck2003pdb}, to a probability that the position is in an IDR.
Following Bludau et al. \cite{bludau2022structural}, the RSA was locally smoothed with a window of $5, 10,15, 20, 25, 30$, or $35$ amino acids, and a sigmoid function was used to predict disorder from this smoothed RSA quantity.
To fit the sigmoid, we used the data in \cite{bludau2022structural} that had IDR labels but no PTM labels.
The smoothing window size used for the final model was the value that resulted in the lowest variance of the bias, $Y - f$, on this data.

\begin{figure}[ht]
    \centering
\includegraphics[width=0.66\textwidth]{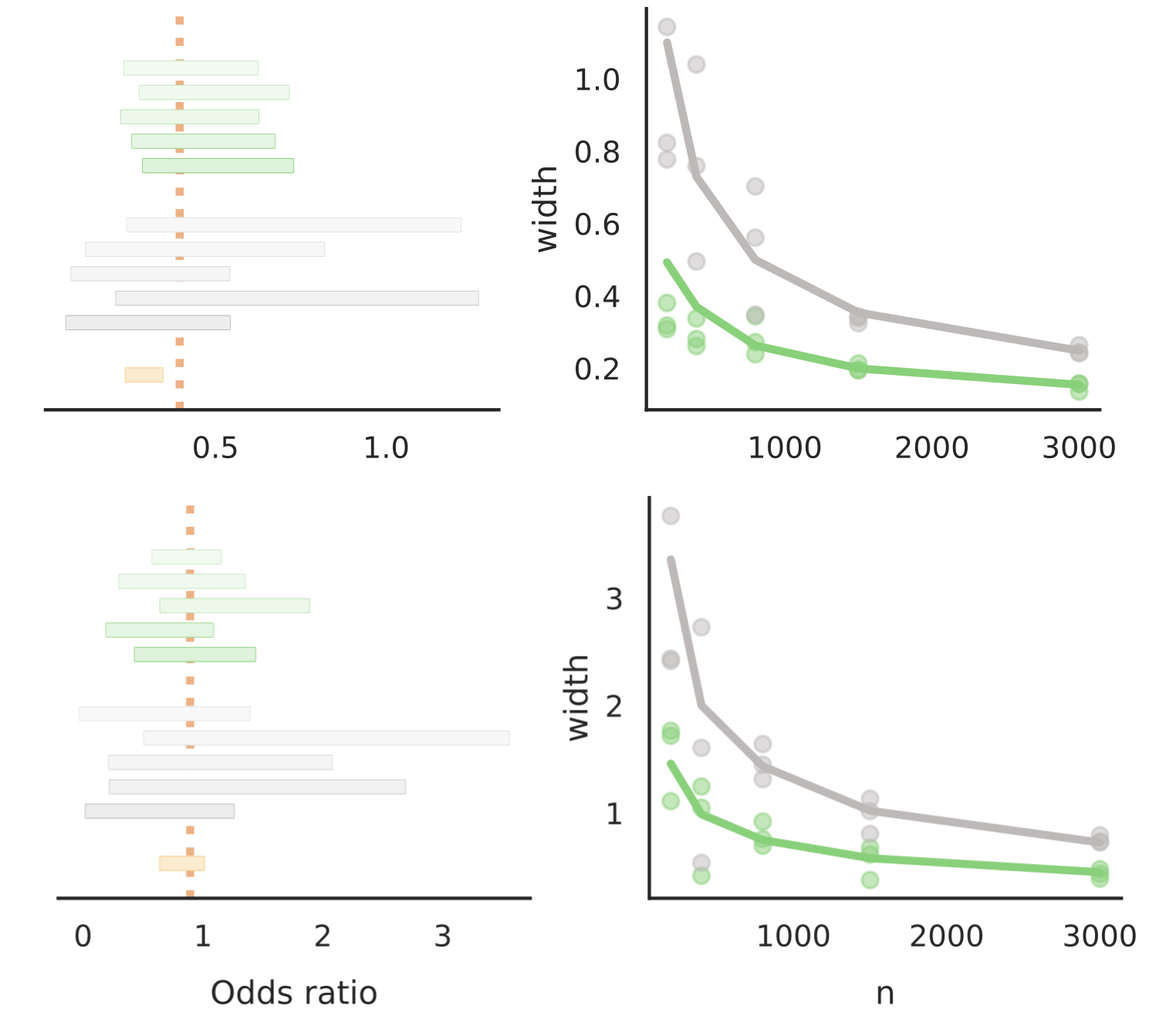}
    \caption{\textbf{Confidence intervals on odds ratio between intrinsic disorder and two types of post-translational modifications}, ubiquitination (top) and acetylation (bottom). Following Figure \ref{fig:big-figure} in the main text, the left panel shows prediction-powered (green) and classical (gray) confidence intervals computed with five random splits of labeled and unlabeled data, as well as the imputation (gold) confidence interval computed using all the unlabeled data. The true value is denoted by the dashed orange line. The right panel shows the average interval width for varying values of $n$, the number of labeled data points, and the width for five randomly chosen trials.}
    \label{fig:odds-ratio-two-types}
\end{figure}

Figure \ref{fig:big-figure}A in the main text presents results on estimating the odds ratio between intrinsic disorder and phosphorylation, a common type of post-translational modification (PTM). Figure \ref{fig:odds-ratio-two-types} shows analogous results on estimating the odds ratio between intrinsic disorder and two other types of PTMs, ubiquitination and acetylation. 
The confidence intervals shown in the left panel of Figure \ref{fig:odds-ratio-two-types} and Figure \ref{fig:big-figure}A in the main text were computed with $n = 400$ labeled data points.

\subsection{Galaxy classification}

We fine-tune a ResNet50 \cite{he2016deep} on the training split of the Galaxy Zoo 2 data with a batch size of 32 and a learning rate of 0.0001 using Adam \cite{kingma2014adam}.
We tune the entire backbone, not just the last layer.
We use the remaining validation split as our labeled and unlabeled data, taking $n\in\{50,100,200,300,500,750,1000\}$.
We use Algorithm \ref{alg:means} for the prediction-powered approach, and Proposition \ref{prop:clt-interval-asymptotic} for the classical and imputation approaches. The confidence intervals shown in the left panel of Figure \ref{fig:big-figure}B in the main text were computed with $n = 366$ labeled data points.

\subsection{Distribution of gene expression levels}
\label{app:genes}

We used the transformer model developed and trained by Vaishnav et al. \cite{vaishnav2022evolution} to predict gene expression level, with the following modification that we found improved predictive performance.
Given $n$ labeled data points, five were randomly selected and used to train an affine (two-parameter) function mapping the scalar prediction of the transformer in \cite{vaishnav2022evolution} to a prediction of the conditional median of the label, using quantile regression.
The predictions of this final model were used for the unlabeled dataset, and the remaining $n - 5$ data points that were not used to fine-tune the transformer model were used as the labeled dataset. The confidence intervals shown in the left panel of Figure \ref{fig:big-figure}C in the main text were computed with $n = 2000$ labeled data points. 

\begin{figure}[ht]
    \centering
\includegraphics[width=0.7\textwidth]{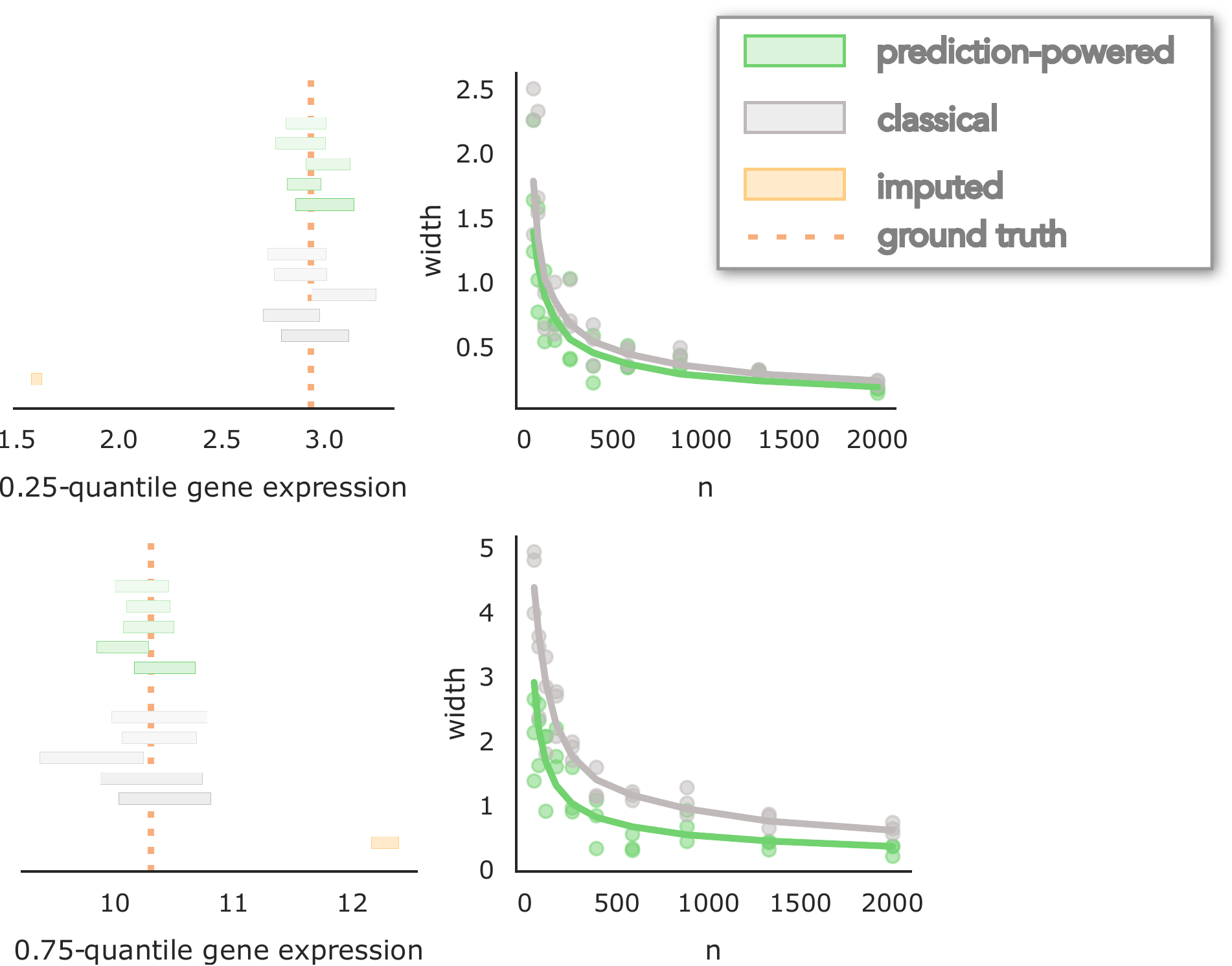}
    \caption{\textbf{Confidence intervals on gene expression quantiles} for $q = 0.25$ (top) and $q = 0.75$ (bottom). Following Figure \ref{fig:big-figure} in the main text, the left panel shows prediction-powered (green) and classical (gray) confidence intervals computed with five random splits of labeled and unlabeled data. The right panel shows the average interval width for varying values of $n$, the number of labeled data points, as well as the width for five randomly chosen trials.}
    \label{fig:gene-expression-quantiles}
\end{figure}

We plot results analogous to Figure \ref{fig:big-figure}C in the main text for the $0.25$- and $0.75$-quantiles in Figure \ref{fig:gene-expression-quantiles}.

\subsection{Estimating deforestation in the Amazon}
\label{app:forest}

The machine-learning model given by \cite{sexton2013global} outputs forest-cover predictions at 30m resolution for 3192 data points.
We correspond these by latitude and longitude with gold-standard data points labeled as one of \texttt{\{deforestation, no deforestation\}} from \cite{bullock2020satellite}.
In the first step, we split off half of the data to train a histogram-based gradient-boosted tree to predict deforestation labels from the forest-cover predictions.
We take a random sample of $n=100$ data points as the gold-standard data, and try to cover the true fraction of deforestation events on the $N=1596$ remaining data points.
We use Algorithm \ref{alg:means} and Proposition \ref{prop:clt-interval-asymptotic} to produce the prediction-powered confidence interval and the classical and imputation intervals, respectively.  The confidence intervals shown in the left panel of Figure \ref{fig:big-figure}D in the main text were computed with $n = 200$ labeled data points.

\subsection{Relationship between income and private health insurance}

We train a gradient-boosted tree \cite{chen2016xgboost} on the California Census data from 2018 acquired using the Folktables \cite{ding2021retiring} interface.
The tree takes as input several covariates such as income, race, and sex, to predict whether an individual has private health insurance coverage.
In the new year, 2019, we use $n\in\{200, 300, 500, 1000, 2000, 5000, 10000\}$ labeled data points. We use Algorithm \ref{alg:logistic} to produce the prediction-powered confidence interval and the standard CLT confidence interval for the classical and imputation approaches. The confidence intervals shown in the left panel of Figure \ref{fig:big-figure}E in the main text were computed with $n = 2000$ labeled data points.

\subsection{Relationship between age and income}

The setting is the same as the above experiment on income and private health insurance, the main difference being that income is used as the target, and not as a covariate.
We used Algorithm \ref{alg:ols} to produce the prediction-powered confidence interval and the standard CLT confidence interval for the classical and imputation approaches. The confidence intervals shown in the left panel of Figure \ref{fig:big-figure}F in the main text were computed with $n = 2000$ labeled data points.

\subsection{Counting plankton}

We fine-tune a ResNet152 \cite{he2016deep} on the WHOI-Plankton dataset \cite{orenstein2015whoi} in the years 2006-2013 for two epochs with a batch size of 32 and a learning rate of 0.0001 using AdamW \cite{loshchilov2017decoupled}, with 5\% of the data saved for validation.
We tune the entire backbone, not just the last layer.
Then we test in the year 2014, using all available data.
We use Theorem \ref{thm:label-shift} to produce the prediction-powered intervals and Proposition \ref{prop:clt-interval-asymptotic} for the imputation approach. The confidence intervals shown in the left panel of Figure \ref{fig:big-figure}G in the main text were computed with $n = 89471$ labeled data points.

\end{document}